\documentclass[sigconf]{acmart}
\AtBeginDocument{%
  }

\acmConference[]{}{}

\usepackage{enumitem}
\setlist[itemize]{leftmargin=10pt, itemsep=1pt}
\usepackage{booktabs}
\usepackage{multirow}
\usepackage[table]{xcolor}

\usepackage{amsmath, amsthm}
\usepackage{bm}
\usepackage{algorithm}
\usepackage{mathtools}
\usepackage{algpseudocode}

\newcommand{\CE}{\mathrm{CE}}
\newcommand{\vs}{\mathbf{s}}
\newcommand{\vm}{\mathbf{m}}
\newcommand{\lse}{\operatorname{lse}}

\renewcommand\footnotetextcopyrightpermission[1]{}
\begin{document}

\title{Select-TTT: Query-Conditioned Selective Fast Weight Updates for In-Place Test-Time Training}

\author{Yue Wu}
\email{wuyuewy@mail.ustc.edu.cn}
\affiliation{%
  \institution{University of Science and Technology of China}
  \country{China}
}

\author{Yang Zhang}
\email{zyang1580@gmail.com}
\affiliation{%
  \institution{National University of Singapore}
  \country{Singapore}
}

\author{Weinan Gan}
\email{ganweinan1@huawei.com}
\affiliation{%
  \institution{Huawei Noah's Ark Lab }
  \country{China}
}

\author{Chongming Gao}
\email{chongminggao@ustc.edu.cn}
\affiliation{%
  \institution{University of Science and Technology of China}
  \country{China}
}

\author{Xinsen Guo}
\email{xsguo@mail.ustc.edu.cn}
\affiliation{%
  \institution{University of Science and Technology of China}
  \country{China}
}

\author{Xiaoyuan Li}
\email{xiaoyuanli@mail.ustc.edu.cn}
\affiliation{%
  \institution{University of Science and Technology of China}
  \country{China}
}

\author{Fuli Feng}
\email{fulifeng93@gmail.com}
\affiliation{%
  \institution{University of Science and Technology of China}
  \country{China}
}

\renewcommand{\shortauthors}{Wu et al.}

\begin{abstract}
Test-time training (TTT) enables a deployed language model to internalize incoming context on-the-fly into a small set of fast-weight parameters, rather than merely treating it as 
transient context for attention. In-Place TTT recently made this idea practical for large 
language models by repurposing the down-projection of standard MLP blocks as fast weights. 
However, we find that its benefit largely disappears on realistic web long-context question 
answering. We attribute this to In-Place TTT's uniform write schedule: every chunk is written 
with equal strength, so irrelevant content dilutes query-relevant information in fast-weight 
memory. Building on this insight, we propose \textbf{Select-TTT}, a model-agnostic, query-conditioned rule for In-Place TTT that scales each chunk's write strength by its relevance to 
the user query. This relevance signal requires no supervision and no new parameters: it is 
read directly from the model's frozen output head, as the probability mass that a chunk's 
hidden states assign to the query tokens. A temperature-scaled softmax turns these scores 
into learning-rate multipliers, leaving the chunk-wise, context-parallel computation of In-Place TTT intact. Select-TTT consistently improves long-context question answering on two 
benchmarks and two backbones. These findings highlight the value of query-conditioned
fast weight writes for practical test-time training on long-context question answering,
which is also potentially promising for long-horizon agent tasks involving multiple user queries 
over evolving contexts.
\end{abstract}





\maketitle

\section{Introduction}

Large Language Models (LLMs)~\cite{brown2020language,openai2024gpt4,llama3,qwen3}
are increasingly deployed as web assistants that answer user queries over
long, evolving contexts: multi-session dialogue histories, web browsing
traces, and lengthy documents~\cite{wu2024longmemeval,locomo}. In such
scenarios, the static ``train then deploy'' paradigm faces a double
challenge. First, the answer-bearing evidence typically occupies only a tiny
fraction of the context, and empirical studies show that LLMs struggle to
exploit long contexts even when they nominally fit the context
window~\cite{liu2023lost,bai2024longbenchv2,hsieh2024ruler,chan2025mlebenchevaluatingmachinelearning}. Second, once
deployed, the model's weights are frozen: the model can re-\emph{read} its
context through attention at every step, but it can never
\emph{internalize} it, leaving its parameters misaligned with the
information at hand~\cite{feng2026inplace}.

Two lines of research address these challenges from complementary angles.
Architectural approaches extend the context window itself, via sparse or
linear
attention~\cite{child2019generating,beltagy2020longformer,zaheer2020bigbird,katharopoulos2020transformers},
state-space
models~\cite{gu2021s4,gu2023mamba,dao2024transformersssmsgeneralizedmodels},
and positional-embedding
scaling~\cite{su2023roformer,yarn}. Test-Time Training
(TTT)~\cite{sun2020ttt,wang2021tent,sun2024tttLayers,behrouz2024titans},
in contrast, makes the model itself adaptive: it designates a small subset
of parameters --- the \emph{fast
weights}~\cite{schmidhuber1992fastweights,fast_weight} ---
as a dynamic memory that is updated on the fly while processing the input,
internalizing contextual information as it arrives. Among TTT methods,
In-Place TTT (IP-TTT)~\cite{feng2026inplace} is of particular practical
interest: it repurposes the final projection matrix of the ubiquitous gated
MLP blocks as fast weights, pairs this with an objective aligned with
Next-Token Prediction (NTP), and organizes updates chunk-wise so that TTT
becomes a ``drop-in'' enhancement compatible with context
parallelism --- requiring no architectural modification and no training from
scratch.

However, the IP-TTT's write schedule is uniform: every chunk updates the fast
weights with the same learning rate, regardless of its content. This is at
odds with how long-context question answering works in practice. When a
question concerns a small portion of a long history, the chunks that carry
the answer are written with the same strength as the many chunks that do
not, and the noise accumulated from irrelevant writes dilutes the
query-relevant information in the fast-weight memory. Empirically, we find
that uniform In-Place TTT brings little or no improvement on realistic
long-context QA benchmarks, and sometimes degrades the zero-shot
performance. A controlled experiment with oracle annotations confirms the
diagnosis: amplifying the writes of answer chunks improves accuracy
markedly, whereas amplifying the writes of irrelevant chunks drives it
below the baseline (Figure~\ref{fig:motivation}).

We solve this question affirmatively with \textbf{Select-TTT}, a
query-conditioned selective update rule for In-Place TTT. Our key
observation is that a pre-trained LLM already carries a well-calibrated
relevance measure --- its own output head. Select-TTT scores each chunk in
the \emph{logit space}: it maps the chunk's hidden states through the frozen
output head and measures the probability mass they assign to the query
tokens (the query-conditioned cross-entropy). A low query-CE indicates that
the local representation is already predictive of the question, i.e., the
chunk is relevant. Chunk-level scores, aggregated by top-$k$ mean pooling,
are converted into per-chunk learning-rate multipliers via a
temperature-scaled softmax designed to meet three desiderata:
\emph{soft writes} (every chunk receives a positive multiplier, so
potentially answer-bearing chunks are never entirely discarded),
\emph{budget preservation} (only the distribution
across chunks changes, conserving
the total amount of information written), and \emph{stability} (an upper
clip prevents any single chunk from dominating). The resulting rule merely
rescales each chunk's update by a scalar, and reduces to IP-TTT when
no query is present.

We evaluate \textbf{Select-TTT} on two long-context QA benchmarks,
LongMemEval-S~\cite{wu2024longmemeval} and LoCoMo~\cite{locomo}, with
Llama-3.1-8B-Instruct and Qwen3-4B backbones, under two protocols:
applying the selective rule only at inference on an In-Place TTT
checkpoint, and enabling it during continual training on
LongCite-45k~\cite{zhang2024longcite} as well. In both settings
Select-TTT consistently outperforms uniform In-Place TTT and the
corresponding baselines. Ablations show that each component of the
multiplier design contributes. 
All these findings highlight the value of query-conditioned fast weight writes for 
practical test-time training.

\section{Preliminary}
\label{sec:prelim}

\subsection{Test-Time Training}
\label{sec:ttt}
Test-Time Training (TTT) enables a model to adapt its parameters at inference
time by introducing a small subset of trainable parameters, the \emph{fast
weights} \cite{fast_weight}. Denoted by $\mathbf{W}$, these weights parameterize a
small network $f_{\mathbf{W}}(\cdot):\mathbb{R}^{d}\to\mathbb{R}^{d}$ that acts
as a dynamic memory, continuously storing and retrieving contextual
information from the input sequence.

Given a sequence of tokens, each token is projected to derive a query
$\mathbf{q}_i$, a key $\mathbf{k}_i$ and a value $\mathbf{v}_i$. The TTT
mechanism alternates two operations:
\begin{enumerate}
  \item \emph{Update Operation:} The fast weight is updated by one optimization
        step of a self-supervised loss $\mathcal{L}(\cdot,\cdot)$ that
        associates the key $\mathbf{k}_i$ with its value $\mathbf{v}_i$:
        \begin{equation}
          \mathbf{W}_{i}\;\leftarrow\;\mathbf{W}_{i-1}
          -\eta\,\nabla_{\mathbf{W}}\,\mathcal{L}\bigl(
            f_{\mathbf{W}_{i-1}}(\mathbf{k}_i),\;\mathbf{v}_i
          \bigr),
          \label{eq:ttt-update}
        \end{equation}
        where $\eta$ is the learning rate. Intuitively, this step encodes the
        information of the pair $(\mathbf{k}_i,\mathbf{v}_i)$ into the memory
        $f_{\mathbf{W}}$.
  \item \emph{Apply Operation:} The updated network is used to process the
        query, i.e., $\mathbf{o}_i = f_{\mathbf{W}_i}(\mathbf{q}_i)$, so that
        the output is enriched with the contextual information stored in
        $\mathbf{W}_i$.
\end{enumerate}

While Eq.~\eqref{eq:ttt-update} describes the high-level mechanism, its
practical instantiation in modern LLMs faces two barriers: (i) TTT is often
realized by \emph{specialized recurrent layers} that replace attention and
require expensive pre-training from scratch; (ii) the canonical
\emph{per-token} update is inherently sequential and severely bottlenecks
parallel accelerators. In-Place TTT \cite{feng2026inplace} addresses both issues, and
serves as the backbone of our method.

\subsection{In-Place TTT}
\label{sec:ipttt}
In-Place TTT \cite{feng2026inplace} repurposes the final projection matrix of the
ubiquitous gated MLP blocks as the fast weight, turning TTT into a
``drop-in'' enhancement without any architectural modification. For the widely used gated MLP
\cite{llama3,qwen3}, given the
hidden representation $\mathbf{H}\in\mathbb{R}^{n\times d_{\mathrm{model}}}$
of $n$ tokens, the gated MLP computes its output as
\begin{equation}
  \mathbf{O}
  =
  \Bigl(
    \phi\bigl(\mathbf{H}\,\mathbf{W}_{\mathrm{gate}}^{\top}\bigr)
    \;\odot\;
    \bigl(\mathbf{H}\,\mathbf{W}_{\mathrm{up}}^{\top}\bigr)
  \Bigr)
  \mathbf{W}_{\mathrm{down}}^{\top},
  \label{eq:gated-mlp}
\end{equation}
where $\phi$ is the activation and $\odot$ the element-wise product. In-Place
TTT freezes $\mathbf{W}_{\mathrm{gate}}$ and $\mathbf{W}_{\mathrm{up}}$ as
\emph{slow weights}, and treats $\mathbf{W}_{\mathrm{down}}$ as the
\emph{fast weight} that is updated at inference time.

\paragraph{LM-aligned objective and chunk-wise update.}
Let
$\mathbf{Z} = \phi(\mathbf{H}\mathbf{W}_{\mathrm{gate}}^{\top})
\odot (\mathbf{H}\mathbf{W}_{\mathrm{up}}^{\top})
\in \mathbb{R}^{n \times d_{\mathrm{ff}}}$
denote the intermediate (gated) activations, and let the value target be
constructed from the token embedding $\mathbf{X}^{0}\in\mathbb{R}^{n\times
d_{\mathrm{model}}}$ so that it carries future-token information,
\begin{equation}
  \widehat{\mathbf{V}}
  \;=\;
  \mathrm{Conv1D}\bigl(\mathbf{X}^{0}\bigr)\,\mathbf{W}_{\mathrm{target}},
  \label{eq:ipttt-target}
\end{equation}
where $\mathrm{Conv1D}(\cdot)$ is a causal 1D convolution and
$\mathbf{W}_{\mathrm{target}}$ is a trainable projection; e.g., a next-token
target is obtained by setting $\mathbf{W}_{\mathrm{target}}$ to the identity
and the convolution kernel to select the next token. In-Place TTT instantiates
the fast-weight loss as the negative Frobenius inner product,
$\mathcal{L}(\cdot,\cdot) = -\langle \cdot,\cdot\rangle_{\mathrm{F}}$,
which is explicitly aligned with the Next-Token-Prediction (NTP) task.

The sequence is partitioned into $T$ non-overlapping chunks of size $C$; for
$\square\in\{\mathbf{Z},\,\widehat{\mathbf{V}},\,\mathbf{O}\}$ we write
$\square_{[i]} = \square_{iC+1:(i+1)C}$ for its $i$-th chunk. Let
$\mathbf{W}^{(i)}_{\mathrm{down}}$ denote the fast-weight state before
processing chunk $i$, with $\mathbf{W}^{(0)}_{\mathrm{down}} =
\mathbf{W}_{\mathrm{down}}$. Processing chunk $i$ performs:
\begin{enumerate}
  \item \emph{Apply Operation:} the current fast weight is used to produce the output
        $\mathbf{O}_{[i]} = \mathbf{Z}_{[i]}
        \bigl(\mathbf{W}^{(i)}_{\mathrm{down}}\bigr)^{\top}$;
  \item \emph{Update Operation:} The fast weight $\mathbf{W}^{(i)}_{\mathrm{down}}$ are updated using $\mathbf{Z}_{[i]}$ as keys and $\widehat{\mathbf{V}}_{[i]}$ as values, which
is performed via one gradient descent step with a loss function  $\mathcal{L}$ and a learning rate $\eta$:

\begin{equation}
  \mathbf{W}^{(i+1)}_{\mathrm{down}}
  \;=\;
  \mathbf{W}^{(i)}_{\mathrm{down}}
  \;-\;
  \eta\,\nabla_{\mathbf{W}}\mathcal{L}
  \left(
    \mathbf{Z}_{[i]}\,(\mathbf{W}^{(i)}_{\mathrm{down}})^{\top},\,
    \widehat{\mathbf{V}}_{[i]}
  \right).
  \label{eq:fast-weight-update}
\end{equation}

Under the loss function $\mathcal{L}(\cdot,\cdot) = -\langle \cdot,\cdot\rangle_{\mathrm{F}}$, the gradient with respect to the fast weights in this chunk-wise mechanism can be directly derived:
        \begin{equation}
          \mathbf{W}^{(i+1)}_{\mathrm{down}}
          \;=\;
          \mathbf{W}^{(i)}_{\mathrm{down}}
          \;+\;
          \eta\,
          \underbrace{\widehat{\mathbf{V}}_{[i]}^{\top}\,\mathbf{Z}_{[i]}}
                   _{\Delta{\mathbf{W}}_{i}},
          \label{eq:ipttt-closed}
        \end{equation}
        with a global learning rate $\eta$ and the chunk update
        $\Delta{\mathbf{W}}_{i} = \widehat{\mathbf{V}}_{[i]}^{\top}\mathbf{Z}_{[i]}$.
\end{enumerate}

The chunk-wise form of Eq.~\eqref{eq:ipttt-closed} keeps the computation dense
and matmul-rich, enabling efficient training with large chunk sizes and
compatibility with context parallelism.

\section{Motivation Experiment}

While In-Place TTT has been shown effective on synthetic retrieval benchmarks
such as RULER~\cite{hsieh2024ruler,feng2026inplace}, 
we observe that its benefit on long-context user QA under realistic web scenarios is
marginal. Concretely, on LongMemEval-S\cite{wu2024longmemeval}, a comprehensive and challenging benchmark for testing the long-term memory of chat assistants, 
enabling In-Place TTT with the Llama-3.1-8B-Instruct backbone raises the average accuracy only from
$33.4\%$ (zero-shot Baseline) to $33.6\%$. We hypothesize that the cause is
the \emph{indiscriminate write} of In-Place TTT: the update of
Eq.~\eqref{eq:ipttt-closed} applies the same learning rate $\eta$ to every
chunk, so chunks irrelevant to the user query are written into the fast
weights with equal strength as the answer-bearing chunks, and their noise
dilutes the query-relevant information in the memory.

To test this hypothesis we design an \emph{ANS-augment} experiment on
Llama-3.1-8B-Instruct. LongMemEval-S provides the session associated with
each question, from which we obtain the ground-truth set of answer-bearing
chunks. We then decouple the write strength of answer chunks from that of
the remaining (irrelevant) chunks and sweep the two learning rates
independently.

\begin{figure}[t]
\centering
\includegraphics[width=0.95\linewidth]{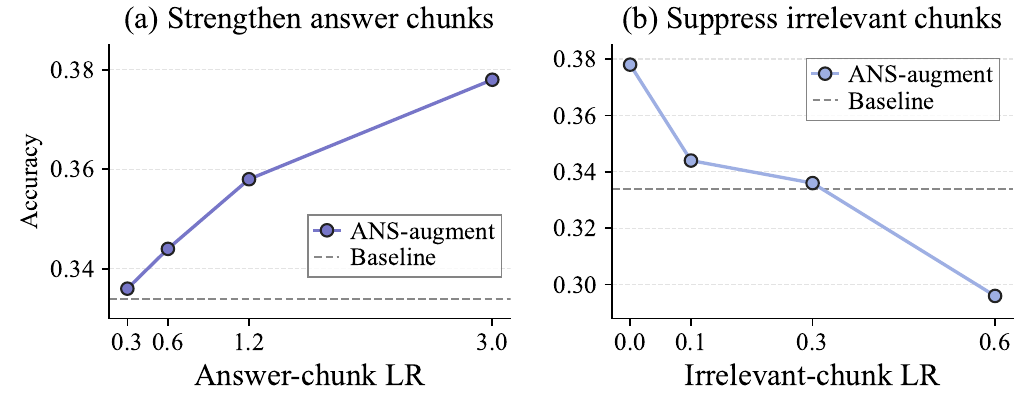}
\caption{Motivation experiment on LongMemEval-S (Llama-3.1-8B-Instruct).
\textbf{(a)}~Accuracy vs.\ the answer-chunk learning rate. \textbf{(b)}~Accuracy vs.\
the irrelevant-chunk learning rate. The default In-Place TTT configuration uses $ 0.3 $ for both learning rates. The dashed line marks the zero-shot Baseline ($33.4\%$).}
\label{fig:motivation}
\end{figure}

Figure~\ref{fig:motivation} reports the results, and two trends are clear.
\emph{(a)} Holding the irrelevant-chunk learning rate fixed, strengthening
the write of answer chunks improves accuracy monotonically, from $33.6\%$
(the uniform In-Place TTT setting) to $37.8\%$. \emph{(b)} Conversely,
increasing the write strength of irrelevant chunks progressively degrades accuracy, from $34 . 8 \% $ down to $29.6 \% $. These two trends reveal a key limitation of uniform
In-Place TTT: the writes from irrelevant chunks actively hurt performance,
while stronger writes to answer chunks help. Then a practical question arises: in realistic 
QA scenarios, how can we know in advance which chunks contain the answer? 
A natural thought is: if each chunk's write strength could be scaled by its relevance to
the query --- amplifying answer-bearing chunks and suppressing irrelevant
ones --- the fast-weight memory would tend to retain the information the model
actually needs. Motivated by this idea, we propose Select-TTT in the next section.

\section{Select-TTT: Query-Conditioned Selective In-Place TTT}
\label{sec:method}

\subsection{The Selective Learning-Rate Rule}
\label{sec:selective-rule}
Our core idea is to modulate the per-chunk memory write by a scalar multiplier
$\lambda_i\in (0, M]$, computed from the relevance between the chunk and the
user query. We observe that this modulation and the In-Place TTT update are
\emph{complementary}: the learning rate sets the \emph{baseline strength} of
each memory write, while the query-conditioned relevance rescales this strength
for each chunk according to how relevant it is to the query. This yields the
\emph{query-conditioned selective learning-rate rule}:

\begin{equation}
  \mathbf{W}^{(i+1)}_{\mathrm{down}}
  \;=\;
  \mathbf{W}^{(i)}_{\mathrm{down}}
  \;+\;
  \underbrace{\eta\,\lambda_i}_{\mathclap{\text{query-conditioned step size}}}\;\widehat{\mathbf{V}}_{[i]}^{\top}\mathbf{Z}_{[i]},
  \label{eq:select-recurrence}
\end{equation}

\begin{figure*}[t]
\centering
\includegraphics[width=\textwidth]{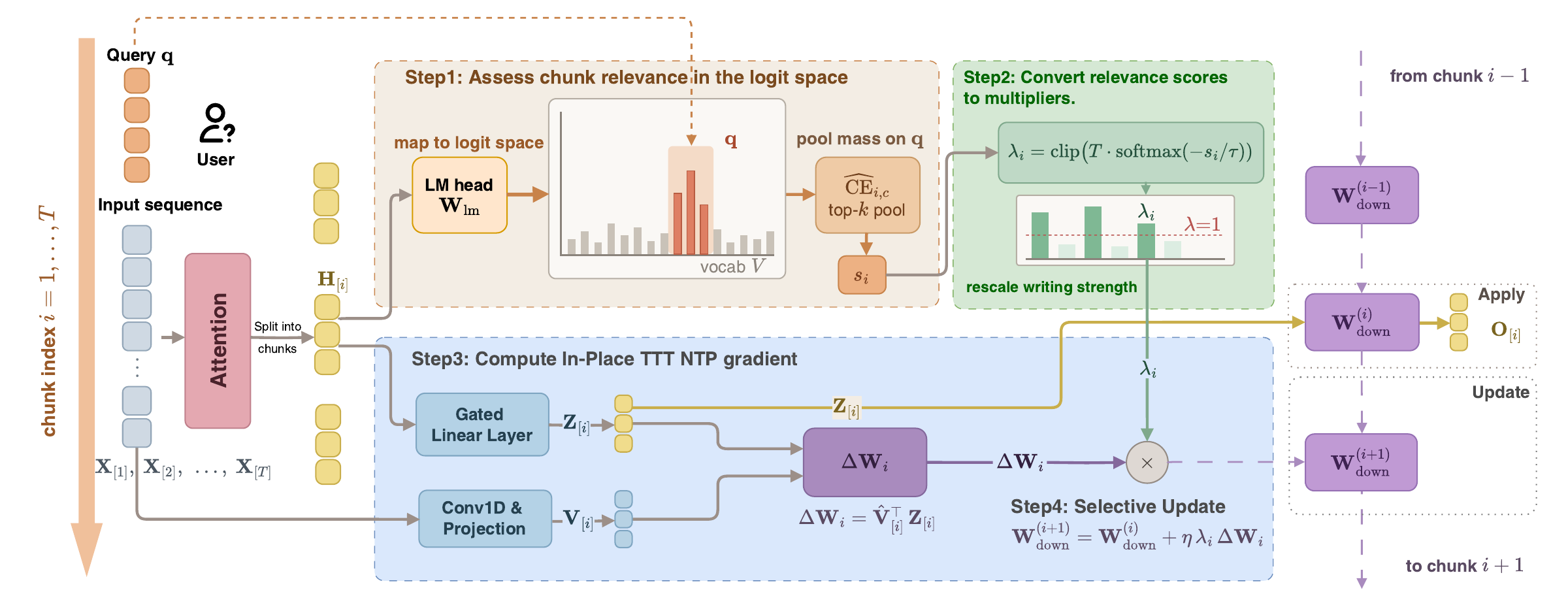}
\caption{Overview of Select-TTT. (1) The frozen LM head maps chunk hidden
states to logit space, where query-token scores are aggregated by top-$k$
mean pooling to assess chunk relevance. (2) Relevance scores are converted
into learning-rate multipliers. (3) In-Place TTT computes the NTP-aligned
fast-weight updates. (4) The multipliers rescale these updates to selectively
write query-relevant information into fast-weight memory.}
\label{fig:framework}
\end{figure*}

This selective rule continuously modulates the memory write strength according to the relevance of each chunk to the query. When a chunk is weakly related to the query, $\lambda_{i}$ becomes small, weakening its memory write so that the fast weight receives only a modest update from that chunk. Conversely, a highly relevant chunk yields $\lambda_{i} > 1$, producing an effective step size $\eta\lambda_{i} > \eta$ and thereby strengthening its write into the memory. Figure~\ref{fig:framework} illustrates our Select-TTT framework.

\subsection{The Query-conditioned chunk selection}
\label{sec:selective-query}
In this subsection, we will detail the procedure by which the query-conditioned multiplier $\lambda_{i}$ is computed. There are two steps:

\paragraph{Assessing chunk relevance score in the logit space.}
A key challenge is that fast weight updates in In-Place TTT are carried out
entirely within the gated MLP, so the only signal available for chunk $i$ is
its own hidden state $\mathbf{H}_{[i]}$, which carries no direct notion of
relevance to the query. Standard retrieval relevance signals such as cosine similarity between a chunk embedding and a query embedding, or a cross-encoder score do not transfer here: they assume documents
are independent units whose vectors come from a \emph{separate},
contrastively-trained encoder that projects query and documents into a shared
calibrated semantic space. The chunks in Select-TTT are instead contiguous
segments of a single input, represented solely by the model's intermediate
hidden states produced online during the forward pass. 

We instead assess the chunk relevance in the \emph{logit space}
induced by the frozen output head $\mathbf{W}_{\mathrm{lm}}$, yielding a novel model-internal mechanism for selecting and
weighting hidden-state chunks when such text-based signals are unavailable: we map each
chunk's hidden state through the output head into a calibrated probability
distribution over the vocabulary, and read off the chunk's importance from the
probability mass it assigns to the query tokens. This can be viewed as projecting each chunk's hidden state 
onto the query-token subspace of the logit space, where the semantics of ``relevance to the query'' 
has a direct, per-token meaning.

Formally, let $\mathbf{q} = (q_{ 1 }, \dots, q_{Q})$ be the query tokens
associated with the current input. During training, $\mathbf{q}$ is the
question associated with each QA instance in the training set; at inference
time, it is provided by the user query. For position $c$ of chunk
$i$, let $\mathbf{h}_{i,c}$ be the hidden state, i.e., the
$c$-th row of $\mathbf{H}_{[i]}$. The output head maps it to the logit vector
\begin{equation}
  \mathbf{z}_{t,c}
  \;=\;
  \mathbf{W}_{\mathrm{lm}}\,
  \mathrm{RMSNorm}\bigl(\mathbf{h}_{i,c}\bigr)
  \;+\;
  \mathbf{b}_{\mathrm{lm}}
  \;\in\;
  \mathbb{R}^{V},
  \label{eq:logits}
\end{equation}
where $V$ is the vocabulary size. The likelihood that the model assigns to the
query tokens at this position is
\begin{equation}
  \widehat{\CE}_{i,c}
  \;=\;
  -\,\frac{1}{Q}
    \sum_{j=1}^{Q}
    \log\,
    \frac{\exp\bigl((\mathbf{z}_{i,c})_{q_j}\bigr)}
         {\sum_{v \in \mathcal{V}} \exp\bigl((\mathbf{z}_{i,c})_v\bigr)}.
  \label{eq:per-token-ce}
\end{equation}
Here $(\mathbf{z}_{i,c})_{q_j}$ refers to the
logit assigned to token $q_j$ at position $c$ of chunk $i$. A low $\widehat{\CE}_{i,c}$ indicates that the local representation already encodes
information predictive of the question, so that the position is relevant to the
query. We aggregate per-position scores within each chunk to obtain a single chunk
relevance score $s_{i}$:

\begin{equation}
  s_i \;=\; \mathrm{Pool}^{(K)}_{c \in [C]}\bigl(\widehat{\CE}_{i,c}\bigr),
  \quad
  K = \max\bigl(1,\ \lceil \rho C \rceil\bigr),
  \label{eq:chunk-ce}
\end{equation}
the $\mathrm{Pool}$ is specified by a strategy governed by the parameter
\emph{top-k ratio} $\rho \in (0,1]$, which controls how many positions within a chunk are retained
for the average:

\begin{itemize}
  \item \emph{mean} ($\rho = 1$): all $C$ positions are pooled, equally weighing
        every token in the chunk;
  \item \emph{max} ($\rho \to 0$): only the single position with the lowest CE
        is retained, giving a most-relevant-token rule;
  \item \emph{top-$k$ mean} ($\rho \in (0,1)$): the $k = \lceil \rho C \rceil$
        positions with the lowest CE are kept and averaged, balancing
        question-relevant signal extraction with robustness to noise.
\end{itemize}

We default to aggregate over positions with a \emph{top-$k$ mean} pooling strategy to obtain a single chunk-level score, since we argue that
the relevance of a chunk should be determined by the importance of its
most query-relevant portions rather than by averaging over all tokens.

\paragraph{Convert relevance scores to multipliers.}
We convert the chunk scores $\vs = [s_1, \dots, s_T]$ into per-chunk
learning-rate multipliers $\vm = [\lambda_1, \dots, \lambda_T]$ via a temperature-scaled
softmax over the negative scores:

\begin{equation}
  \lambda_t \;=\; \mathrm{clip}\!\left(
    \frac{\exp(-s_t / \tau)}{\sum_{i=1}^{T} \exp(-s_i / \tau)} \cdot T,
    \; 0, \; M
  \right).
  \label{eq:multiplier}
\end{equation}
Here the \emph{temperature} $\tau $ controls the extent
to which query-relevant chunks are distinguished from irrelevant ones in the
multiplier assignment. A smaller $\tau $ sharpens this distinction: it amplifies
the gap between relevant and irrelevant chunks, concentrating the write budget
on the few most relevant chunks. The clip $M$ sets the
maximum allowed amplification per chunk.
The design of Eq.~\eqref{eq:multiplier} satisfies three desiderata:

\begin{enumerate}
  \item \textbf{Soft write.} The softmax yields strictly positive
    multipliers for \emph{all} chunks, so every chunk is written into the
    fast weights with a strength proportional to its estimated relevance.
    Since the query-CE assessment is only an estimate, the smooth
    allocation ensures that the potential answer-bearing chunks are never
    entirely ignored.
  \item \textbf{Budget preservation.} Multiplying the softmax weights by $T$
    ensures $\frac{1}{T}\sum_{t=1}^{T} \lambda_t \approx 1$, so the average
    learning rate matches IP-TTT's global $\eta$. The total amount of
    information written into fast weights is conserved; only its distribution
    across chunks changes.
  \item \textbf{Stability.} The upper clip $M$ prevents any single chunk from
    dominating the update when the temperature $\tau$ is small, which guards against any individual chunk
being written too aggressively into the fast weights.
\end{enumerate}


\paragraph{An online-learning view.}
Interpreting $\Delta{\mathbf{W}}_{i}$ of Eq.~\eqref{eq:ipttt-closed} as a closed-form gradient
step of the NTP-aligned objective \cite{feng2026inplace}, Eq.~\eqref{eq:select-recurrence}
is exactly chunk-wise SGD with a \emph{query-conditioned, per-chunk adaptive
step size} $\eta\lambda_t$: the overall write strength is set by the global learning rate $\eta$, then
rescaled by the query-conditioned multiplier $\lambda_t$
before the write is performed. Per-block adaptive step
sizes are a classical technique in deep learning optimization, and here the
step size is conditioned on the query, so that the model adaptively focuses
its memory on query-relevant content. 

\subsection{Efficient Chunk-Wise Scoring and Training}
\label{sec:efficient}
The selective rule only \emph{rescales} each $\Delta{\mathbf{W}}_{i}$ by a per-chunk scalar
$\lambda_i$. The main chunk-wise pass of In-Place TTT
(Eq.~\eqref{eq:ipttt-closed}) is therefore left
untouched, and all of its matmul-rich, context-parallel properties are
preserved. The only additional computation is the scoring pass of
Eqs.~\eqref{eq:logits}--\eqref{eq:chunk-ce}, which we make hardware-efficient
with two techniques.

\paragraph{Blockwise log-sum-exp.}
The logits in Eq.~\eqref{eq:logits} span the full vocabulary $V$ and would
materialize a tensor of size $[T, C, V]$ if computed directly, incurring
substantial GPU memory overhead. We instead compute the log-sum-exp normalizer
in vocabulary blocks $\{B_k\}_{k}$ of a smaller size $B$,
\begin{equation}
  \begin{aligned}
    \lse_{i,c}
    &=
    \log\sum_{v \in \mathcal{V}} \exp\bigl((\mathbf{z}_{i,c})_v\bigr) \\
    &=
    \log\sum_{k}
    \exp\Bigl(
      \underbrace{
        \log\sum_{v \in B_k} \exp\bigl((\mathbf{z}_{i,c})_v\bigr)
      }_{\lse^{(k)}_{i,c}}
    \Bigr),
  \end{aligned}
  \label{eq:blockwise-lse}
\end{equation}
where each block-wise $\lse^{(k)}_{i,c}$ is a single matrix product against a
block of the (frozen) output head, and the blocks are combined by
$\log\!\text{-}\!\exp$ additions. This keeps the peak activation memory at
$O\bigl(B \cdot T \cdot C\bigr)$ instead of $O\bigl(V \cdot T \cdot C\bigr)$.

\paragraph{Direct question-token logits.}
Only the $Q$ rows of the output head that correspond to the question tokens are
needed in Eq.~\eqref{eq:per-token-ce}. We gather these rows and compute the
logits directly,
\begin{equation}
  \begin{aligned}
    (\mathbf{z}_{i,c})_{q_j}
    &=
    \mathbf{w}^{\top}_{\mathrm{lm},q_j}\,\mathrm{RMSNorm}\bigl(\mathbf{h}_{i,c}\bigr)
    \;+\;
    b_{\mathrm{lm},q_j},
    && j = 1,\dots,Q,
  \end{aligned}
  \label{eq:question-logits}
\end{equation}
with cost $O\bigl(T \cdot C \cdot H \cdot Q\bigr)$ instead of the
$O\bigl(T \cdot C \cdot H \cdot V\bigr)$ of a full forward pass. For question
lengths $Q \ll V$ this is substantially cheaper. Expanding the
logarithm in Eq.~\eqref{eq:per-token-ce}, the per-token CE
decomposes as
\begin{equation}
  \begin{aligned}
    \widehat{\CE}_{i,c}
    &=
    -\,\frac{1}{Q}
      \sum_{j=1}^{Q}
      \log\,
      \frac{\exp\bigl((\mathbf{z}_{i,c})_{q_j}\bigr)}
           {\sum_{v \in \mathcal{V}} \exp\bigl((\mathbf{z}_{i,c})_v\bigr)} \\
    &=
    -\,\frac{1}{Q}
      \sum_{j=1}^{Q}
      \Bigl[
        (\mathbf{z}_{i,c})_{q_j}
        -
        \log\sum_{v \in \mathcal{V}} \exp\bigl((\mathbf{z}_{i,c})_v\bigr)
      \Bigr] \\
    &=
    \frac{1}{Q}
      \sum_{j=1}^{Q}
      \bigl[\, \lse_{i,c} - (\mathbf{z}_{i,c})_{q_j} \,\bigr],
  \end{aligned}
  \label{eq:ce-decomposed}
\end{equation}
where the normalizer $\lse_{i,c}$ is obtained from Eq.~\eqref{eq:blockwise-lse}
and the gathered logits $(\mathbf{z}_{i,c})_{q_j}$ from
Eq.~\eqref{eq:question-logits}. The per-token CE therefore decomposes into two
cheap terms --- $Q$ gathered logits plus a single blockwise log-sum-exp ---
requiring no $[T, C, V]$ logit tensor.

\paragraph{Overhead Analysis.}
Per selective layer, the scoring pass adds
$O\bigl(T C H (V + Q)\bigr)$ floating-point operations --- dominated by the
blockwise log-sum-exp, i.e., a single matrix product against the output head
--- and $O\bigl(T C (B + Q)\bigr)$ peak activation memory. For each chunk, the additional computational cost amounts a single
output-head forward pass over the input sequence per selective layer. The
output head is frozen and shared across layers, so no additional parameters or
optimizer states are introduced.
Since this selective mechanism is executed only at the TTT layers
and is computed from hidden states that are already available, the added
overhead is modest relative to the overall per-step computation of the model.

\paragraph{Efficient Implementation with Context Parallelism.}
Our implementation inherits the context parallelism from In-Place TTT, 
which is designed for high efficiency and scalability on modern hardware.
The full algorithm is presented in Algorithm~\ref{alg:select-ttt} (Appendix~\ref{app:algorithm}).

\section{Experiments}
\label{sec:exp}

In this section, we organize our empirical study around the following research questions:
\begin{itemize}
  \item \textbf{Q1 (Effectiveness).} When the Select-TTT is \emph{only} enabled during
        inference, as a training-free, drop-in enhancement for the In-Place TTT,
        does query-conditioned selectivity improve the zero-shot long-context QA
        performance?
  \item \textbf{Q2 (Training-inference consistency).} When the Select-TTT is enabled during
        \emph{both} continual training and inference, does it outperform
        standard continual pre-training and uniformly trained In-Place TTT?
  \item \textbf{Q3 (Sensitivity).} How sensitive is our Select-TTT framework to its key design choices?
  \item \textbf{Q4 (Mechanism).} Does the query-conditioned selective
        mechanism of Eq.~\eqref{eq:chunk-ce} genuinely rank query-relevant chunks above
        irrelevant ones?
\end{itemize}

\subsection{Experimental Setup}
\label{sec:setup}

\paragraph{Datasets.}
We evaluate our method on two widely adopted long-context QA
benchmarks that cover complementary aspects of long-term memory.
\textbf{LongMemEval-S}~\cite{wu2024longmemeval} focuses on interactive memory in
multi-session user-assistant dialogues, where information is accumulated across
sessions and may evolve over time. 
It contains 500 questions embedded in
histories of approximately 40 sessions and 115K tokens each, and comprises six
subtasks.
\textbf{LoCoMo}~\cite{locomo}, in contrast, tests reasoning over very long
conversations that span up to 35 sessions and roughly 300 turns per dialogue,
with a total of around 1,980 questions distributed across 10 dialogues.
Together, these benchmarks allow us to assess both
multi-session consistency and long-context retrieval under a unified setup.
For continual-training experiments, we use \textbf{LongCite-45k}~\cite{zhang2024longcite},
a large-scale corpus of 44,600 long-context QA instances (up to 128K tokens)
which provides a natural source of
query-context pairs for training.

\paragraph{Backbone models.}
We use two backbones of different families and scales:
\textit{Llama-3.1-8B-Instruct} ($32$ layers) and \textit{Qwen3-4B} ($36$
layers). For the \textit{Qwen3-4B} model, we adapt the model’s Rotary Position Embeddings using YaRN~\cite{yarn} to effectively manage these long sequences.
Following the In-Place TTT configuration, every sixth layer starting
from the first one (layer indices $\{0, 6, 12, 18, 24, 30\}$, i.e., six layers
per backbone) is equipped with the TTT mechanism; all other weights are
identical to the original checkpoints.

\paragraph{Evaluation protocols.}
We consider two complementary protocols.
\emph{Protocol I (inference-time Select-TTT).} The Select-TTT chunk-wise updates of Eq.~\eqref{eq:select-recurrence}
are applied only at inference-time. We compare: (i) \textbf{Baseline} --- the
original pre-trained LLM evaluated directly, without any test-time update;
(ii) \textbf{IP-TTT} --- In-Place TTT with a uniform learning rate, i.e.,
$\lambda_i = 1$ for all chunks $i$; and (iii) \textbf{Select-TTT} --- our
query-conditioned multipliers (Eq.~\eqref{eq:multiplier}). Methods (ii) and
(iii) are run on the \emph{same} TTT-enhanced checkpoint, obtained by following the original
recipe of In-Place TTT; the two rows differ \emph{only} in whether the
per-chunk multiplier is uniform or query-conditioned, so their contrast
cleanly isolates the effect of query-conditioned selective mechanism.
\emph{Protocol II (selective continual training).} The model is continually
trained for up to $5$k steps on the processed LongCite-45k corpus with the
TTT mechanism active during training, and evaluated on LongMemEval-S at
every 1k-step checkpoint with the same mechanism active at inference. We
compare: (i) \textbf{CPT} --- standard continual pre-training on the same
corpus without any TTT mechanism; (ii) \textbf{IP-TTT} --- continual training
with uniform TTT updates ($\lambda_i = 1$); and (iii) \textbf{Select-TTT}
--- continual training with the selective mechanism. For TTT-based training, both the backbone parameters and the newly 
introduced TTT parameters are updated jointly so that the model adapts its TTT behavior to the long-context QA
setting, whereas CPT follows the vanilla fine-tuning recipe. We also report
the \textbf{No-Training} reference, i.e., the zero-shot performance before
any continual training.

\paragraph{Implementation details.}
All TTT-based methods use the LM-aligned objective of In-Place TTT. In the
continual-training protocol we set chunk size $C = 2048$ and TTT learning
rate $\eta = 0.3$. Select-TTT defaults to temperature $\tau = 0.1$, cap
$M = 10.0$, pooling ratio $\rho = 0.1$, and top-$k$ mean pooling in
Eq.~\eqref{eq:chunk-ce}. The ablation
(Section~\ref{sec:ablation}) varies $\tau$, $\rho$, and the pooling strategy
around these defaults.
Detailed descriptions of the experiment implementation can be found in
Appendix~\ref{app:implementation}.

\begin{table}[t]
\centering
\caption{Protocol I: zero-shot inference-time TTT on LongMemEval-S with two
backbones. Accuracy (\%) is reported; the best result within each backbone
block is in \textbf{bold}.}
\label{tab:infer-longmemeval}
\footnotesize
\setlength{\tabcolsep}{3pt}
\begin{tabular}{cc cccccc|c}
\toprule
\multirow{2}{*}{\textbf{Backbone}}
& \multirow{2}{*}{\textbf{Method}}
& \multicolumn{7}{c}{\textbf{LongMemEval-S}}
\\
\cmidrule(lr){3-9}
& & \textbf{KU} & \textbf{MS} & \textbf{TR}
& \textbf{S-Pref.} & \textbf{S-Asst.} & \textbf{S-User} & \textbf{Avg.}
\\
\midrule
\multirow{3}{*}{\textit{Llama-3.1-8B}}
& Baseline & 50.00 & 18.80 & 16.54 & 3.33 & \textbf{69.64} & \textbf{58.57} & 33.4
\\
& IP-TTT & 47.44 & 18.80 & 20.30 & 3.33 & 66.07 & \textbf{58.57} & 33.6
\\
\rowcolor[HTML]{F2F2F2}
& \textbf{Select-TTT} & \textbf{52.56} & \textbf{22.56} & \textbf{28.57}
& \textbf{13.33} & 64.29 & 54.29 & \textbf{37.4}
\\
\midrule
\multirow{3}{*}{\textit{Qwen3-4B}}
& Baseline & 28.21 & 21.80 & \textbf{22.56} & \textbf{6.67} & 57.14 & 30.00 & 27.2
\\
& IP-TTT & 26.92 & 15.79 & 20.30 & 3.33 & \textbf{67.86} & 27.14 & 25.4
\\
\rowcolor[HTML]{F2F2F2}
& \textbf{Select-TTT} & \textbf{29.49} & \textbf{26.32} & \textbf{22.56}
& 3.33 & 57.14 & \textbf{32.86} & \textbf{28.8}
\\
\bottomrule
\end{tabular}
\vspace{1mm}

{\scriptsize KU: Knowledge Update; MS: Multi-session; TR: Temporal
Reasoning; S-Pref.: Single-session Preference; S-Asst.: Single-session
Assistant; S-User: Single-session User.}
\end{table}

\begin{table}[t]
\centering
\caption{Protocol I: zero-shot inference-time TTT on LoCoMo with two
backbones. Overall score (\%) is reported; the best result within each backbone
block is in \textbf{bold}.}
\label{tab:infer-locomo}
\footnotesize
\setlength{\tabcolsep}{3pt}
\begin{tabular}{cc ccccc|c}
\toprule
\multirow{2}{*}{\textbf{Backbone}}
& \multirow{2}{*}{\textbf{Method}}
& \multicolumn{6}{c}{\textbf{LoCoMo}}
\\
\cmidrule(lr){3-8}
& & \textbf{Multi} & \textbf{Temp.} & \textbf{Open}
& \textbf{Single} & \textbf{Adv.} & \textbf{Avg.}
\\
\midrule
\multirow{3}{*}{\textit{Llama-3.1-8B}}
& Baseline & \textbf{36.60} & 29.30 & 17.78 & 57.91 & 18.83 & 39.55
\\
& IP-TTT & 34.92 & 30.12 & 19.51 & 58.22 & 15.47 & 38.90
\\
\rowcolor[HTML]{F2F2F2}
& \textbf{Select-TTT} & 36.38 & \textbf{30.74} & \textbf{20.74}
& \textbf{58.34} & \textbf{20.40} & \textbf{40.42}
\\
\midrule
\multirow{3}{*}{\textit{Qwen3-4B}}
& Baseline & 25.66 & 20.12 & 18.31 & 38.88 & 4.04 & 25.15
\\
& IP-TTT & \textbf{25.84} & 20.60 & 18.19 & 39.25 & 4.26 & 25.45
\\
\rowcolor[HTML]{F2F2F2}
& \textbf{Select-TTT} & 25.27 & \textbf{21.12} & \textbf{20.46}
& \textbf{39.77} & \textbf{6.05} & \textbf{26.19}
\\
\bottomrule
\end{tabular}
\vspace{1mm}

{\scriptsize Multi: Multi-hop; Temp.: Temporal reasoning; Open: Open-domain
factual; Single: Single-hop short-memory; Adv.: Adversarial queries.}
\end{table}

\subsection{Inference-Time Select-TTT as a Drop-in Enhancement}
\label{sec:inference}
This section addresses \textbf{Q1} under Protocol I: the TTT-enhanced checkpoint is obtained by 
the original recipe of In-Place TTT and the only adaptation at test time is the
selective chunk-wise updates of Eq.~\eqref{eq:select-recurrence}.
Tables~\ref{tab:infer-longmemeval} and \ref{tab:infer-locomo} report the
zero-shot results on LongMemEval-S and LoCoMo for both backbones.

\paragraph{Results.}
Select-TTT attains the best average score on both benchmarks and both
backbones: on LongMemEval-S it improves over IP-TTT by $+3.8$ (Llama) and
$+3.4$ (Qwen), and over the Baseline by $+4.0$ and $+1.6$ respectively; on
LoCoMo it reaches $40.42$ (Llama) and $26.19$ (Qwen), again the highest among
the three methods. More revealing is the \emph{relative} behavior of the two
TTT variants. The uniform adaptation of IP-TTT does not consistently help in
this zero-shot long-context QA scenario: on Qwen3-4B it \emph{drops} the LongMemEval-S
average from $27.2$ to $25.4$, and on Llama-3.1-8B-Instruct from $39.55$ to $38.90$. 
This indicates that In-Place TTT writes \emph {every} chunk into the fast weights with \emph {equal} 
strength dilutes query-relevant information, as irrelevant chunks contribute noise on equal
footing with relevant ones. Simply replacing
the uniform schedule with query-conditioned multipliers on the
\emph{same} checkpoint yields a consistent gain: by suppressing
the write strength of query-irrelevant chunks while reinforcing that of
relevant ones, Select-TTT achieves the best performance across both backbones and both benchmarks.

A further observation is that the Select-TTT gains
concentrate on subtasks that require synthesizing an answer from multiple,
scattered pieces of evidence: on
LongMemEval-S, the Knowledge Update, Multi-session, and Temporal Reasoning show larger improvements
compare to the Single-session subtasks. This indicates that Select-TTT selectively strengthens the writes of query-relevant chunks into the fast weights, which is particularly beneficial for multi-evidence reasoning that aggregates information from multiple chunks.
These observations answer
\textbf{Q1} affirmatively: as a drop-in replacement for the uniform
schedule, the Select-TTT turns out to be a
consistent, training-free enhancement for the In-Place TTT.

\subsection{Continual Pre-Training with Select-TTT}
\label{sec:training}
\begin{figure}[t]
\centering
\includegraphics[width=0.95\linewidth]{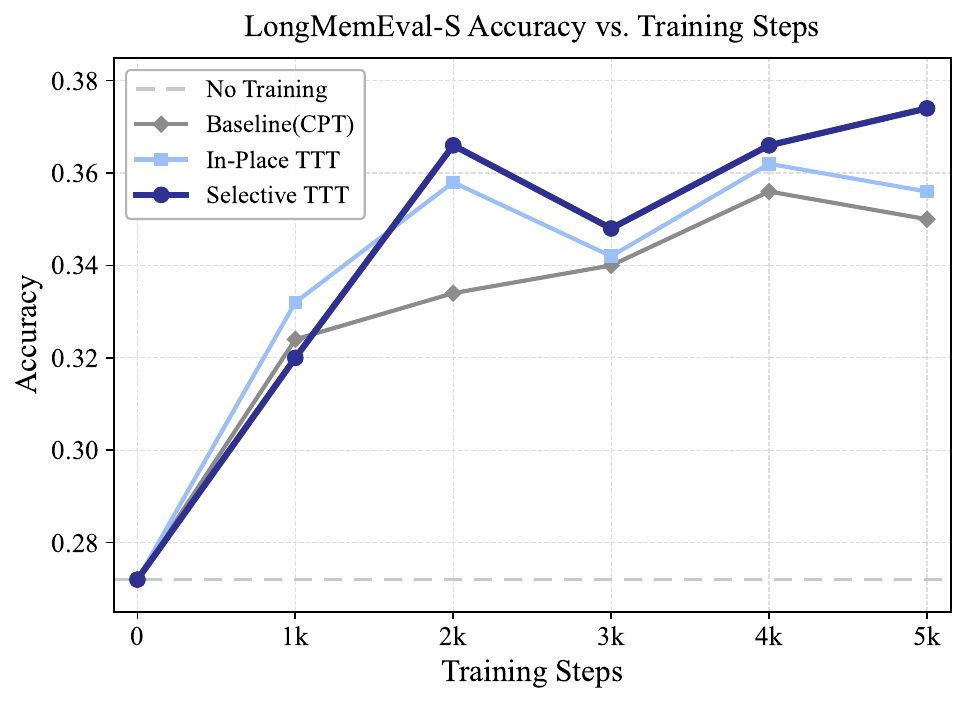}
\caption{Protocol II: LongMemEval-S overall accuracy vs.\ continual-training
steps for No Training, CPT, IP-TTT, and Select-TTT (Qwen3-4B backbone).
}
\label{fig:steps}
\end{figure}

We next address \textbf{Q2} under Protocol II, where the TTT mechanism is
active during both continual training and inference. All methods are trained
on the processed LongCite-45k corpus for up to $5$k steps with the Qwen3-4B
backbone and evaluated on LongMemEval-S at every 1k-step checkpoint.
Figure~\ref{fig:steps} plots the overall accuracy during the continual-training
steps.


\paragraph{Results.}
Continual training on LongCite-45k transfers well: every method lifts the
zero-shot reference of $27.2\%$ above $32\%$ within the first thousand
steps. The comparison among the trained models is more informative. From
step 2k onward, both TTT-based methods outperform standard CPT, confirming
that the In-Place TTT mechanism is an effective vehicle for long-context
adaptation when backbone and TTT parameters are trained jointly; and
Select-TTT in turn matches or exceeds IP-TTT at every subsequent checkpoint,
finishing first at $37.4\%$, ahead of IP-TTT ($35.6\%$) and CPT
($35.0\%$). The single exception occurs at the earliest checkpoint, where
the model has barely begun to accommodate the TTT mechanism and the
query-conditioned redistribution of the learning-rate budget is not yet
beneficial; as training proceeds, the selective schedule consistently
dominates. Because Select-TTT and IP-TTT share the same update rule and
differ \emph{only} in the per-chunk multiplier $\lambda_i$ of
Eq.~\eqref{eq:multiplier}, the persistent gap between the two curves
isolates the contribution of selectivity and answers \textbf{Q2}
affirmatively. 
Full per-subtask results and additional results across all checkpoints are provided
in the Appendix~\ref{app:cpt-results}.

\subsection{Ablation Studies On the Impact of Select-TTT Design Choices}
\label{sec:ablation}
\begin{figure*}[t]
\centering
\includegraphics[width=0.95\textwidth]{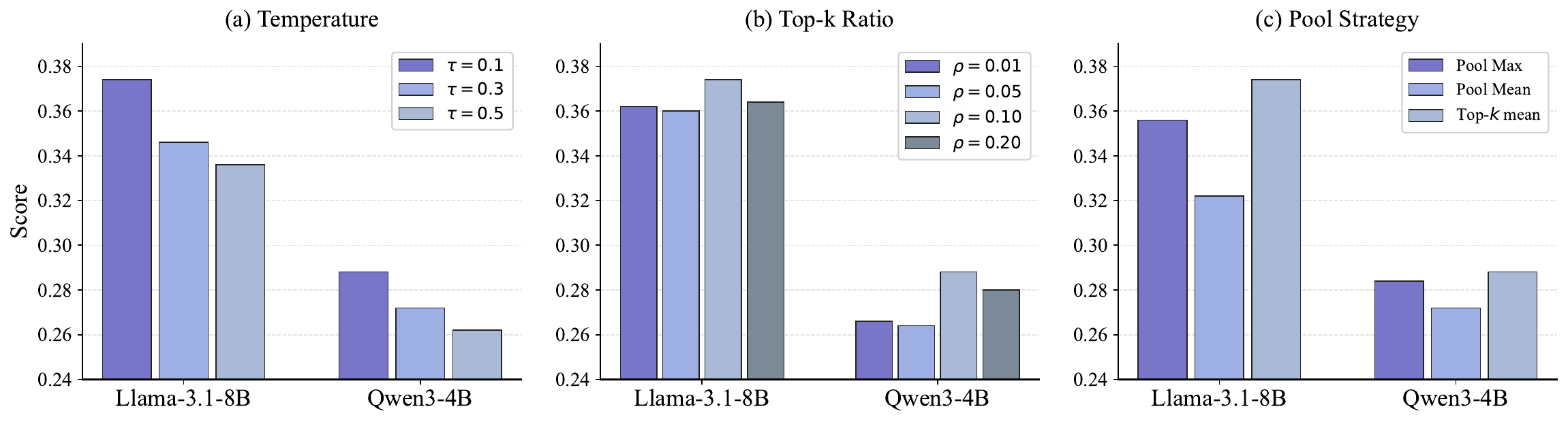}
\caption{Ablation of Select-TTT on LongMemEval-S (both backbones):
(a) temperature $\tau \in \{0.1, 0.3, 0.5\}$; (b) pooling ratio
$\rho \in \{0.01, 0.05, 0.10, 0.20\}$; (c) pooling strategy (Max, Mean,
Top-$k$). The default configuration is $\tau=0.1$, $\rho=0.10$, top-$k$
mean.}
\label{fig:ablation}
\end{figure*}

To answer \textbf{Q3}, we ablate the key design choices of Select-TTT under
Protocol I: (a) the temperature $\tau$ in Eq.~\eqref{eq:multiplier}, which
controls how sharply the learning-rate budget concentrates on low-CE
chunks; (b) the pooling ratio $\rho$ in Eq.~\eqref{eq:chunk-ce}; (c) the
pooling strategy that aggregates per-token query-CE into a chunk-level
score; and (d) the two safeguards of the multiplier --- the softmax,
which realizes a soft write over all chunks, and the upper clip $M$, which
prevents over-writes. Figure~\ref{fig:ablation} reports (a)--(c) on both
backbones with the LongMemEval-S overall accuracy;
Table~\ref{tab:ablation-mult} reports (d) with the Llama-3.1-8B backbone
on all six subtasks.

\paragraph{Temperature $\tau$.}
Figure~\ref{fig:ablation} (a) shows that sharper selection helps: the lowest temperature $\tau = 0.1$
performs best on both backbones, and performance degrades as $\tau$
increases. A small $\tau$ tends to concentrate the
learning-rate budget on the few most query-relevant chunks. This aligns
with the motivation of Section~\ref{sec:selective-rule}: what harms the fast
weights is precisely the dilution caused by spreading the budget over
irrelevant chunks.

\paragraph{Pooling ratio $\rho$.}
Unlike the temperature, the overall accuracy varies non-monotonically with the pooling ratio \(\rho\); the
accuracy varies across the different settings, reflecting an inherent trade-off in Figure~\ref{fig:ablation} (b). 
We settle on $\rho = 0.10$ as a compromise: enough positions participate to make the
score robust against spurious matches, while it still concentrates on the
most query-relevant content.

\begin{table}[t]
\centering
\caption{Ablation of the multiplier design (Eq.~\eqref{eq:multiplier}) on
LongMemEval-S with the Llama-3.1-8B backbone. Accuracy (\%) is
reported; the best result per column is in \textbf{bold}.}
\label{tab:ablation-mult}
\footnotesize
\setlength{\tabcolsep}{3pt}
\begin{tabular}{c cccccc|c}
\toprule
\multirow{2}{*}{\textbf{Ablation Setting}}
& \multicolumn{7}{c}{\textbf{LongMemEval-S}}
\\
\cmidrule(lr){2-8}
& \textbf{KU} & \textbf{MS} & \textbf{TR}
& \textbf{S-Pref.} & \textbf{S-Asst.} & \textbf{S-User} & \textbf{Avg.}
\\
\midrule
\rowcolor[HTML]{F2F2F2}
w/ soft write, w/ clip & \textbf{52.56} & \textbf{22.56} & \textbf{28.57}
& \textbf{13.33} & 64.29 & 54.29 & \textbf{37.4}
\\
w/ soft write, w/o clip & 46.15 & 21.05 & 23.31 & 3.33 & \textbf{66.07}
& \textbf{57.14} & 34.6
\\
w/o soft write, w/ clip & 42.31 & 15.04 & 18.05 & 6.67 & \textbf{66.07}
& 51.43 & 30.4
\\
w/o soft write, w/o clip & 21.79 & 13.53 & 17.29 & 3.33 & 53.57 & 37.14
& 23.0
\\
\bottomrule
\end{tabular}
\end{table}

\begin{table}[t]
\centering
\caption{Comparing relevance signals across different chunk sizes on LongMemEval-S QA instances.
\textbf{Hit@$k$}: at least one of the $k$ highest-ranked chunks contains the
answer. \textbf{Rec@$k$}: fraction of answer chunks covered by the $k$
highest-ranked chunks.}
\label{tab:rationality}
\footnotesize
\setlength{\tabcolsep}{3.5pt}
\begin{tabular}{c c ccc ccc}
\toprule
\multirow{2}{*}{\textbf{Chunk size}} & \multirow{2}{*}{\textbf{Signal}}
& \multicolumn{3}{c}{\textbf{Hit@$k$}}
& \multicolumn{3}{c}{\textbf{Rec@$k$}}
\\
\cmidrule(lr){3-5}\cmidrule(lr){6-8}
& & $k{=}1$ & $k{=}3$ & $k{=}5$ & $k{=}1$ & $k{=}3$ & $k{=}5$
\\
\midrule
\rowcolor[HTML]{F2F2F2}
\multirow{3}{*}{2048}
& \textbf{Query-CE (ours)} & \textbf{0.200} & \textbf{0.440} & \textbf{0.582} & \textbf{0.092} & \textbf{0.201} & \textbf{0.278}
\\
& Cosine & 0.094 & 0.212 & 0.293 & 0.061 & 0.136 & 0.189
\\
& Cross-encoder & 0.090 & 0.245 & 0.372 & 0.020 & 0.059 & 0.098
\\
\midrule
\rowcolor[HTML]{F2F2F2}
\multirow{3}{*}{4096}
& \textbf{Query-CE (ours)} & \textbf{0.256} & \textbf{0.519} & \textbf{0.654} & \textbf{0.128} & \textbf{0.295} & \textbf{0.393}
\\
& Cosine & 0.133 & 0.308 & 0.415 & 0.085 & 0.198 & 0.267
\\
& Cross-encoder & 0.126 & 0.334 & 0.493 & 0.040 & 0.119 & 0.199
\\
\midrule
\rowcolor[HTML]{F2F2F2}
\multirow{3}{*}{8192}
& \textbf{Query-CE (ours)} & \textbf{0.312} & \textbf{0.653} & \textbf{0.843} & \textbf{0.183} & \textbf{0.381} & \textbf{0.599}
\\
& Cosine & 0.189 & 0.398 & 0.522 & 0.122 & 0.257 & 0.337
\\
& Cross-encoder & 0.195 & 0.486 & 0.684 & 0.082 & 0.247 & 0.312
\\
\bottomrule
\end{tabular}
\end{table}

\paragraph{Pooling strategy.}
Figure~\ref{fig:ablation} (c) shows that the \emph{top-$k$ mean} pooling strategy outperforms both \emph{Max} and \emph{Mean} pooling 
on both backbones, validating our hypothesis in Section~\ref{sec:selective-query} that a
chunk's relevance should be governed by its most query-relevant portions.
\emph{Mean} is diluted by the many irrelevant positions within a chunk, 
while \emph{Max} relies on a single position and inherits its fragility. \emph{Top-$k$ mean} strikes the
right balance: it aggregates a small, robust subset of the most relevant
positions, capturing strong query-relevant signal while remaining robust
to individual noisy tokens.

\paragraph{Multiplier design.}
A final ablation examines the two safeguards built into the multiplier of
Eq.~\eqref{eq:multiplier} (Table~\ref{tab:ablation-mult}). The softmax
realizes a \emph{soft write}: the query-CE ranking is only an estimate of
relevance, so writing \emph{every} chunk with a strength proportional to its
estimated relevance ensures that answer-bearing chunks are never entirely
skipped. The clip, in turn, guards against over-writes: an extremely
confident chunk would otherwise attract a multiplier far beyond the average
budget and be written into the fast weights too aggressively. 
The result in Table~\ref{tab:ablation-mult} shows that removing both 
components causes a performance drop.
Both safeguards are empirically necessary. 
\subsection{Analysis: Which Relevance Signal Picks the Relevant Chunks?}
\label{sec:analysis}
The preceding sections show \emph{that} selectivity helps; here we answer
\textbf{Q4} and examine why. To rule out the possibility that a
standard retrieval signal would serve equally well despite the mismatch
argued in Section~\ref{sec:selective-query}, we compare the query-CE score
of Eq.~\eqref{eq:chunk-ce} with two representative alternatives:
\textbf{Cosine} similarity between the mean-pooled chunk and
query hidden states; and a pretrained \textbf{Cross-encoder} bge-reranker~\cite{bge-reranker}
consuming the hidden states.
On LongMemEval-S QA instances with
ground-truth answer-chunk annotations, we rank the chunks under each
signal across different chunk sizes and measure \textbf{Hit@$k$} --- whether at least one of the $k$
highest-ranked chunks contains the answer and \textbf{Rec@$k$} ---
the fraction of answer chunks they cover.
\paragraph{Results.}
Table~\ref{tab:rationality} shows that query-CE is the strongest signal at
every chunk size and $k$ under both metrics. This is consistent with our
argument in Section~\ref{sec:selective-query}: under this hidden-state input setting, query-CE
provides a stronger chunk-ranking signal than cosine similarity or the
pretrained reranker which are not aligned with the model's
hidden states. These results answer \textbf{Q4}: the logit-space query-CE score turns out to be an effective relevance signal.

\subsection{Overhead analysis}
 
\begin{figure}[t]
\centering
\includegraphics[width=0.95\linewidth]{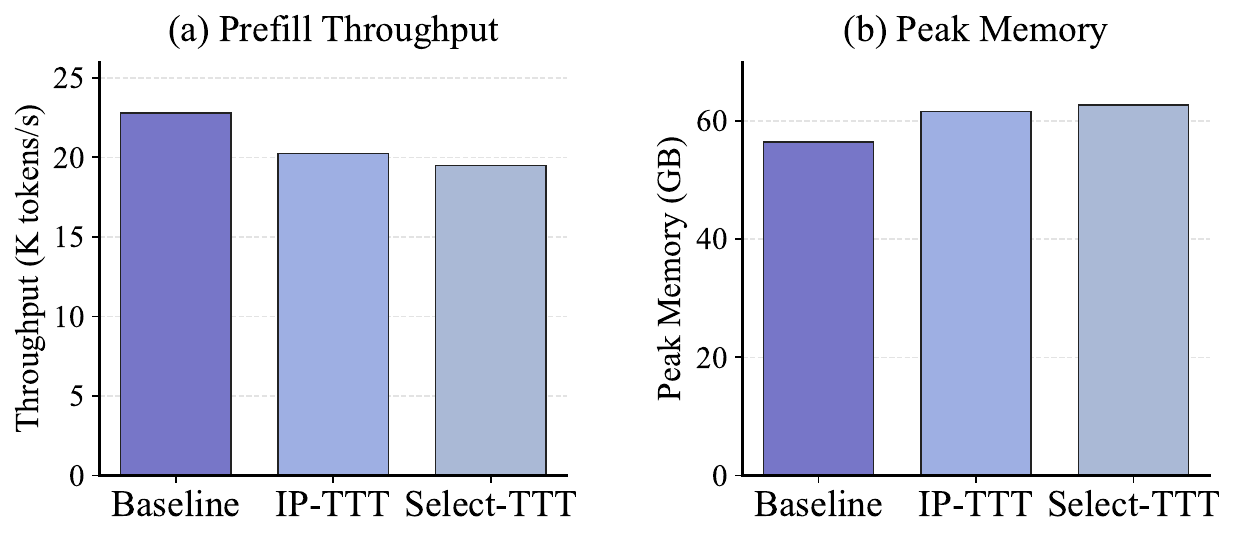}
\caption{Efficiency analysis of Select-TTT. Prefill throughput (a) and peak
GPU memory (b) are compared across the Baseline, IP-TTT, and
Select-TTT on Qwen3-4B model.}
\label{fig:efficiency}
\end{figure}

\label{sec:efficiency}
We finally assess the computational overhead of Select-TTT relative to
IP-TTT and the no-TTT baseline on Qwen3-4B model during inference time. Figure~\ref{fig:efficiency} 
reports the prefill throughput and peak GPU memory measured on different 
settings, confirming that the Select-TTT's lightweight scoring mechanism 
introduces negligible overhead to the original IP-TTT.

\section{Related Work}
\label{sec:related}
\paragraph{Test-Time Training and fast weights.}
Fast weights, network parameters updated on the fly to act as a dynamic
memory, date back to the early
nineties~\cite{schmidhuber1992fastweights,fast_weight} and were
revived by linear
transformers~\cite{fast_weight,katharopoulos2020transformers}. Test-time
training was introduced in computer vision as a means of adapting to
distribution shift~\cite{sun2020ttt,wang2021tent} and has since been
scaled to language modeling, both as standalone sequence
models~\cite{sun2024tttLayers,behrouz2024titans,karami2025lattice,zhang2025tttdr}
and, most recently, as an in-place enhancement of pre-trained
LLMs~\cite{feng2026inplace}. Research has concentrated on the update
objective, the optimizer, and the parallelization of the inherently
sequential per-token
updates~\cite{sun2023retentive,irie2025fastweightprogramminglinear,yau2025sequentialparalleldualityprefixscannable}.
In all of these methods, however, the write schedule is fixed in advance:
every token or chunk is written with the same strength, independently of
its relevance to the task the model is being asked to perform.

\paragraph{Efficient long-context modeling.}
A parallel line of work attacks the quadratic cost of
attention~\cite{vaswani2017attention} directly, through sparse attention
patterns~\cite{child2019generating,beltagy2020longformer,zaheer2020bigbird},
streaming attention with sinks~\cite{xiao2023efficient}, IO-aware exact
attention~\cite{dao2022flashattention}, linear
attention~\cite{katharopoulos2020transformers,fast_weight}, and
state-space
models~\cite{gu2021s4,gu2023mamba,dao2024transformersssmsgeneralizedmodels,yang2024delta},
complemented by positional-embedding scaling~\cite{su2023roformer,yarn}
and segment-level recurrence~\cite{dai2019transformerxl}. These methods
make long inputs processable, yet benchmark studies show that the context
models actually exploit falls well short of the advertised
window~\cite{liu2023lost,bai2024longbenchv2,hsieh2024ruler}. Test-time
adaptation is orthogonal to this line: rather than processing the context
more cheaply, it changes what the model retains from the context, and the
two can be combined.

\paragraph{Memory and gating in sequence models.}
Equipping language models with memory has a long history, from retrieval
over external
corpora~\cite{khandelwal2019knnlm,guu2020realm,lewis2020rag} and
dedicated memory
modules~\cite{graves2014ntm,raecompressive2019,dai2019transformerxl} to
recent memory systems for conversational
assistants~\cite{packer2023memgpt,zhong2023memorybank,chhikara2025mem0,wang2024memoryllmselfupdatablelargelanguage,yu2025memagent}. These systems keep memory outside the model's
parameters and decide what to read at answer time, typically through
similarity retrieval. A complementary mechanism operates inside
linear-recurrent models, where data-dependent gates control what a
parametric state retains~\cite{gu2023mamba,yang2025gated}.

\section{Conclusion}
\label{sec:conclusion}
We introduced Select-TTT, which applies a query-conditioned selective
learning-rate rule that scores each chunk in the logit space of the frozen
output head and rescales its fast-weight write accordingly. 
The rule is
a drop-in enhancement that preserves the chunk-wise, context-parallel
computation of IP-TTT and adds only a lightweight scoring pass. 
This provides
a novel model-internal way to select hidden states when
conventional text-based retrieval signals are unavailable or poorly matched.
Across two
backbones and two benchmarks, Select-TTT consistently improves over IP-TTT
and zero-shot baselines, both as an inference-time enhancement
and when coupled with continual training. We see Select-TTT as a step toward query-aware test-time memory, and
expect the principle --- conditioning what is written on why it is being
written --- to extend to other test-time adaptation mechanisms and to
agentic web applications where long, evolving contexts are the norm.

\section{Limitations}
\label{sec:limitations}
Select-TTT conditions its memory writes on the user query, and the
logit-space query-CE is only a heuristic estimate of chunk relevance: 
the
scoring rule applies the frozen LM head to states from multiple
TTT layers, 
potentially affecting the
calibration of their query-token scores.
A learned relevance function is a promising direction but may introduce additional
computation, parameters, and training overhead. 
Our fixed chunk size follows the IP-TTT setting
to preserve context parallelism. Semantically adaptive chunking could yield
better boundaries and relevance estimates.
Our evaluation covers
two backbones of up to 8B parameters and two English benchmarks; scaling to
larger models, multilingual settings, and other modalities remains future
work.

\bibliographystyle{ACM-Reference-Format}
\bibliography{sample-base}

@InProceedings{sun2020ttt,
  title = 	 {Test-Time Training with Self-Supervision for Generalization under Distribution Shifts},
  author =       {Sun, Yu and Wang, Xiaolong and Liu, Zhuang and Miller, John and Efros, Alexei and Hardt, Moritz},
  booktitle = 	 {Proceedings of the 37th International Conference on Machine Learning},
  pages = 	 {9229--9248},
  year = 	 {2020},
  editor = 	 {III, Hal Daumé and Singh, Aarti},
  volume = 	 {119},
  series = 	 {Proceedings of Machine Learning Research},
  month = 	 {13--18 Jul},
  publisher =    {PMLR},
}

@inproceedings{fast_weight,
author = {Ba, Jimmy and Hinton, Geoffrey and Mnih, Volodymyr and Leibo, Joel Z. and Ionescu, Catalin},
title = {Using fast weights to attend to the recent past},
year = {2016},
isbn = {9781510838819},
publisher = {Curran Associates Inc.},
address = {Red Hook, NY, USA},
booktitle = {Proceedings of the 30th International Conference on Neural Information Processing Systems},
pages = {4338–4346},
numpages = {9},
location = {Barcelona, Spain},
series = {NIPS'16}
}

@article{llama3,
  title={The llama 3 herd of models},
  author={Grattafiori, Aaron and Dubey, Abhimanyu and Jauhri, Abhinav and Pandey, Abhinav and Kadian, Abhishek and Al-Dahle, Ahmad and Letman, Aiesha and Mathur, Akhil and Schelten, Alan and Vaughan, Alex and others},
  journal={arXiv preprint arXiv:2407.21783},
  year={2024}
}

@misc{qwen3,
      title={Qwen3 Technical Report}, 
      author={An Yang and Anfeng Li and Baosong Yang and Beichen Zhang and Binyuan Hui and Bo Zheng and Bowen Yu and Chang Gao and Chengen Huang and Chenxu Lv and others},
      year={2025},
      eprint={2505.09388},
      archivePrefix={arXiv},
      primaryClass={cs.CL},
      url={https://arxiv.org/abs/2505.09388}, 
}

@article{wu2024longmemeval,
      title={LongMemEval: Benchmarking Chat Assistants on Long-Term Interactive Memory}, 
      author={Di Wu and Hongwei Wang and Wenhao Yu and Yuwei Zhang and Kai-Wei Chang and Dong Yu},
      year={2024},
      eprint={2410.10813},
      archivePrefix={arXiv},
      primaryClass={cs.CL},
      url={https://arxiv.org/abs/2410.10813}, 
}

@misc{bge-reranker,
      title={C-Pack: Packed Resources For General Chinese Embeddings}, 
      author={Shitao Xiao and Zheng Liu and Peitian Zhang and Niklas Muennighoff and Defu Lian and Jian-Yun Nie},
      year={2024},
      eprint={2309.07597},
      archivePrefix={arXiv},
      primaryClass={cs.CL},
      url={https://arxiv.org/abs/2309.07597}, 
}

@inproceedings{zhang2024longcite,
    title = "{L}ong{C}ite: Enabling {LLM}s to Generate Fine-grained Citations in Long-Context {QA}",
    author = "Zhang, Jiajie  and
      Bai, Yushi  and
      Lv, Xin  and
      Gu, Wanjun  and
      Liu, Danqing  and
      Zou, Minhao  and
      Cao, Shulin  and
      Hou, Lei  and
      Dong, Yuxiao  and
      Feng, Ling  and
      Li, Juanzi",
    booktitle = "Findings of the {A}ssociation for {C}omputational {L}inguistics: {ACL} 2025",
    month = jul,
    year = "2025",
    address = "Vienna, Austria",
    publisher = "Association for Computational Linguistics",
    url = "https://aclanthology.org/2025.findings-acl.264/",
    doi = "10.18653/v1/2025.findings-acl.264",
    pages = "5098--5122",
    ISBN = "979-8-89176-256-5",
}

@inproceedings{yarn,
 author = {Peng, Bowen and Quesnelle, Jeffrey and Fan, Honglu and Shippole, Enrico},
 booktitle = {International Conference on Learning Representations},
 editor = {B. Kim and Y. Yue and S. Chaudhuri and K. Fragkiadaki and M. Khan and Y. Sun},
 pages = {31932--31951},
 title = {YaRN: Efficient Context Window Extension of Large Language Models},
 volume = {2024},
 year = {2024}
}

@article{hsieh2024ruler,
  title={RULER: What's the Real Context Size of Your Long-Context Language Models?},
  author={Cheng-Ping Hsieh and Simeng Sun and Samuel Kriman and Shantanu Acharya and Dima Rekesh and Fei Jia and Yang Zhang and Boris Ginsburg},
  year={2024},
  journal={arXiv preprint arXiv:2404.06654},
}

@inproceedings{brown2020language,
  title={Language models are few-shot learners},
  author={Brown, Tom and Mann, Benjamin and Ryder, Nick and Subbiah, Melanie and Kaplan, Jared D and Dhariwal, Prafulla and Neelakantan, Arvind and Shyam, Pranav and Sastry, Girish and Askell, Amanda and others},
  booktitle = {Advances in Neural Information Processing Systems},
  volume={33},
  pages={1877--1901},
  year={2020}
}

@misc{openai2024gpt4,
      title={GPT-4 Technical Report}, 
      author={OpenAI and Josh Achiam and Steven Adler and Sandhini Agarwal and Lama Ahmad and Ilge Akkaya and Florencia Leoni Aleman and Diogo Almeida and Janko Altenschmidt and Sam Altman and others},
      year={2024},
      eprint={2303.08774},
      archivePrefix={arXiv},
      primaryClass={cs.CL},
      url={https://arxiv.org/abs/2303.08774}, 
}

@ARTICLE{schmidhuber1992fastweights,
  author={Schmidhuber, Jürgen},
  journal={Neural Computation}, 
  title={Learning to Control Fast-Weight Memories: An Alternative to Dynamic Recurrent Networks}, 
  year={1992},
  volume={4},
  number={1},
  pages={131-139},
  doi={10.1162/neco.1992.4.1.131}}

@misc{wang2021tent,
      title={Tent: Fully Test-time Adaptation by Entropy Minimization}, 
      author={Dequan Wang and Evan Shelhamer and Shaoteng Liu and Bruno Olshausen and Trevor Darrell},
      year={2021},
      eprint={2006.10726},
      archivePrefix={arXiv},
      primaryClass={cs.LG},
      url={https://arxiv.org/abs/2006.10726}, 
}

@InProceedings{sun2024tttLayers,
  title = 	 {Learning to ({L}earn at Test Time): {RNN}s with Expressive Hidden States},
  author =       {Sun, Yu and Li, Xinhao and Dalal, Karan and Xu, Jiarui and Vikram, Arjun and Zhang, Genghan and Dubois, Yann and Chen, Xinlei and Wang, Xiaolong and Koyejo, Sanmi and Hashimoto, Tatsunori and Guestrin, Carlos},
  booktitle = 	 {Proceedings of the 42nd International Conference on Machine Learning},
  pages = 	 {57503--57522},
  year = 	 {2025},
  volume = 	 {267},
  series = 	 {Proceedings of Machine Learning Research},
  month = 	 {13--19 Jul},
  publisher =    {PMLR},
}

@inproceedings{behrouz2024titans,
 author = {Behrouz, Ali and Zhong, Peilin and Mirrokni, Vahab},
 booktitle = {Advances in Neural Information Processing Systems},
 doi = {10.52202/085713-3786},
 editor = {D. Belgrave and C. Zhang and H. Lin and R. Pascanu and P. Koniusz and M. Ghassemi and N. Chen},
 pages = {113506--113543},
 publisher = {Curran Associates, Inc.},
 title = {Titans: Learning to Memorize at Test Time},
 url = {https://proceedings.neurips.cc/paper_files/paper/2025/file/a4ca07aa108036f80cbb5b82285fd4b1-Paper-Conference.pdf},
 volume = {38, Main Conference},
 year = {2025}
}

@misc{karami2025lattice,
      title={Lattice: Learning to Efficiently Compress the Memory}, 
      author={Mahdi Karami and Razvan Pascanu and Vahab Mirrokni},
      year={2025},
      eprint={2504.05646},
      archivePrefix={arXiv},
      primaryClass={cs.LG},
      url={https://arxiv.org/abs/2504.05646}, 
}

@misc{zhang2025tttdr,
      title={Test-Time Training Done Right}, 
      author={Tianyuan Zhang and Sai Bi and Yicong Hong and Kai Zhang and Fujun Luan and Songlin Yang and Kalyan Sunkavalli and William T. Freeman and Hao Tan},
      year={2025},
      eprint={2505.23884},
      archivePrefix={arXiv},
      primaryClass={cs.LG},
      url={https://arxiv.org/abs/2505.23884}, 
}

@inproceedings{feng2026inplace,
  title={In-place test-time training},
  author={Feng, Guhao and Luo, Shengjie and Hua, Kai and Zhang, Ge and Huang, Wenhao and He, Di and Cai, Tianle},
  booktitle={International Conference on Learning Representations},
  volume={2026},
  pages={114452--114472},
  year={2026}
}

@misc{sun2023retentive,
      title={Retentive Network: A Successor to Transformer for Large Language Models}, 
      author={Yutao Sun and Li Dong and Shaohan Huang and Shuming Ma and Yuqing Xia and Jilong Xue and Jianyong Wang and Furu Wei},
      year={2023},
      eprint={2307.08621},
      archivePrefix={arXiv},
      primaryClass={cs.CL},
      url={https://arxiv.org/abs/2307.08621}, 
}

@misc{irie2025fastweightprogramminglinear,
      title={Fast weight programming and linear transformers: from machine learning to neurobiology}, 
      author={Kazuki Irie and Samuel J. Gershman},
      year={2026},
      eprint={2508.08435},
      archivePrefix={arXiv},
      primaryClass={cs.LG},
      url={https://arxiv.org/abs/2508.08435}, 
}

@misc{yau2025sequentialparalleldualityprefixscannable,
      title={Sequential-Parallel Duality in Prefix Scannable Models}, 
      author={Morris Yau and Sharut Gupta and Valerie Engelmayer and Kazuki Irie and Stefanie Jegelka and Jacob Andreas},
      year={2026},
      eprint={2506.10918},
      archivePrefix={arXiv},
      primaryClass={cs.LG},
      url={https://arxiv.org/abs/2506.10918}, 
}

@inproceedings{vaswani2017attention,
 author = {Vaswani, Ashish and Shazeer, Noam and Parmar, Niki and Uszkoreit, Jakob and Jones, Llion and Gomez, Aidan N and Kaiser, Lukasz and Polosukhin, Illia},
 booktitle = {Advances in Neural Information Processing Systems},
 pages = {},
 publisher = {Curran Associates, Inc.},
 title = {Attention is All you Need},
 volume = {30},
 year = {2017}
}

@misc{su2023roformer,
      title={RoFormer: Enhanced Transformer with Rotary Position Embedding}, 
      author={Jianlin Su and Yu Lu and Shengfeng Pan and Ahmed Murtadha and Bo Wen and Yunfeng Liu},
      year={2023},
      eprint={2104.09864},
      archivePrefix={arXiv},
      primaryClass={cs.CL},
      url={https://arxiv.org/abs/2104.09864}, 
}

@misc{child2019generating,
      title={Generating Long Sequences with Sparse Transformers}, 
      author={Rewon Child and Scott Gray and Alec Radford and Ilya Sutskever},
      year={2019},
      eprint={1904.10509},
      archivePrefix={arXiv},
      primaryClass={cs.LG},
      url={https://arxiv.org/abs/1904.10509}, 
}

@misc{beltagy2020longformer,
      title={Longformer: The Long-Document Transformer}, 
      author={Iz Beltagy and Matthew E. Peters and Arman Cohan},
      year={2020},
      eprint={2004.05150},
      archivePrefix={arXiv},
      primaryClass={cs.CL},
      url={https://arxiv.org/abs/2004.05150}, 
}

@inproceedings{zaheer2020bigbird,
 author = {Zaheer, Manzil and Guruganesh, Guru and Dubey, Kumar Avinava and Ainslie, Joshua and Alberti, Chris and Ontanon, Santiago and Pham, Philip and Ravula, Anirudh and Wang, Qifan and Yang, Li and Ahmed, Amr},
 booktitle = {Advances in Neural Information Processing Systems},
 editor = {H. Larochelle and M. Ranzato and R. Hadsell and M.F. Balcan and H. Lin},
 pages = {17283--17297},
 publisher = {Curran Associates, Inc.},
 title = {Big Bird: Transformers for Longer Sequences},
 volume = {33},
 year = {2020}
}

@misc{xiao2023efficient,
      title={Efficient Streaming Language Models with Attention Sinks}, 
      author={Guangxuan Xiao and Yuandong Tian and Beidi Chen and Song Han and Mike Lewis},
      year={2024},
      eprint={2309.17453},
      archivePrefix={arXiv},
      primaryClass={cs.CL},
      url={https://arxiv.org/abs/2309.17453}, 
}

@inproceedings{dao2022flashattention,
 author = {Dao, Tri and Fu, Dan and Ermon, Stefano and Rudra, Atri and R\'{e}, Christopher},
 booktitle = {Advances in Neural Information Processing Systems},
 doi = {10.52202/068431-1189},
 editor = {S. Koyejo and S. Mohamed and A. Agarwal and D. Belgrave and K. Cho and A. Oh},
 pages = {16344--16359},
 publisher = {Curran Associates, Inc.},
 title = {FlashAttention: Fast and Memory-Efficient Exact Attention with IO-Awareness},
 url = {https://proceedings.neurips.cc/paper_files/paper/2022/file/67d57c32e20fd0a7a302cb81d36e40d5-Paper-Conference.pdf},
 volume = {35},
 year = {2022}
}

@inproceedings{katharopoulos2020transformers,
author = {Katharopoulos, Angelos and Vyas, Apoorv and Pappas, Nikolaos and Fleuret, Fran{\c c}ois},
title = {Transformers are RNNs: fast autoregressive transformers with linear attention},
year = {2020},
publisher = {JMLR.org},
booktitle = {Proceedings of the 37th International Conference on Machine Learning},
articleno = {478},
numpages = {10},
series = {ICML'20}
}

@inproceedings{gu2021s4,
  title     = {{Efficiently Modeling Long Sequences with Structured State Spaces}},
  author    = {Gu, Albert and Goel, Karan and Re, Christopher},
  booktitle = {International Conference on Learning Representations},
  year      = {2022},
  url       = {https://mlanthology.org/iclr/2022/gu2022iclr-efficiently/}
}

@misc{gu2023mamba,
      title={Mamba: Linear-Time Sequence Modeling with Selective State Spaces}, 
      author={Albert Gu and Tri Dao},
      year={2024},
      eprint={2312.00752},
      archivePrefix={arXiv},
      primaryClass={cs.LG},
      url={https://arxiv.org/abs/2312.00752}, 
}

@inproceedings{dao2024transformersssmsgeneralizedmodels,
author = {Dao, Tri and Gu, Albert},
title = {Transformers are SSMs: generalized models and efficient algorithms through structured state space duality},
year = {2024},
publisher = {JMLR.org},
booktitle = {Proceedings of the 41st International Conference on Machine Learning},
articleno = {399},
numpages = {31},
location = {Vienna, Austria},
series = {ICML'24}
}

@inproceedings{yang2024delta,
 author = {Yang, Songlin and Wang, Bailin and Zhang, Yu and Shen, Yikang and Kim, Yoon},
 booktitle = {Advances in Neural Information Processing Systems},
 doi = {10.52202/079017-3668},
 pages = {115491--115522},
 publisher = {Curran Associates, Inc.},
 title = {Parallelizing Linear Transformers with the Delta Rule over Sequence Length},
 volume = {37},
 year = {2024}
}

@inproceedings{dai2019transformerxl,
  title        = {Transformer-{XL}: Attentive Language Models Beyond a Fixed-Length Context},
  author       = {Dai, Zihang and Yang, Zhilin and Yang, Yiming and Carbonell, Jaime and Le, Quoc V. and Salakhutdinov, Ruslan},
  booktitle    = {Proceedings of the 57th Annual Meeting of the Association for Computational Linguistics},
  pages        = {2978--2988},
  year         = {2019},
  publisher    = {Association for Computational Linguistics},
  url          = {https://aclanthology.org/P19-1285/}
}

@article{liu2023lost,
    title = "Lost in the Middle: How Language Models Use Long Contexts",
    author = "Liu, Nelson F.  and
      Lin, Kevin  and
      Hewitt, John  and
      Paranjape, Ashwin  and
      Bevilacqua, Michele  and
      Petroni, Fabio  and
      Liang, Percy",
    journal = "Transactions of the Association for Computational Linguistics",
    volume = "12",
    year = "2024",
    address = "Cambridge, MA",
    publisher = "MIT Press",
    url = "https://aclanthology.org/2024.tacl-1.9/",
    doi = "10.1162/tacl_a_00638",
    pages = "157--173",
}

@misc{bai2024longbenchv2,
      title={LongBench v2: Towards Deeper Understanding and Reasoning on Realistic Long-context Multitasks}, 
      author={Yushi Bai and Shangqing Tu and Jiajie Zhang and Hao Peng and Xiaozhi Wang and Xin Lv and Shulin Cao and Jiazheng Xu and Lei Hou and Yuxiao Dong and Jie Tang and Juanzi Li},
      year={2025},
      eprint={2412.15204},
      archivePrefix={arXiv},
      primaryClass={cs.CL},
      url={https://arxiv.org/abs/2412.15204}, 
}

@misc{khandelwal2019knnlm,
      title={Generalization through Memorization: Nearest Neighbor Language Models}, 
      author={Urvashi Khandelwal and Omer Levy and Dan Jurafsky and Luke Zettlemoyer and Mike Lewis},
      year={2020},
      eprint={1911.00172},
      archivePrefix={arXiv},
      primaryClass={cs.CL},
      url={https://arxiv.org/abs/1911.00172}, 
}

@inproceedings{guu2020realm,
author = {Guu, Kelvin and Lee, Kenton and Tung, Zora and Pasupat, Panupong and Chang, Ming-Wei},
title = {REALM: retrieval-augmented language model pre-training},
year = {2020},
publisher = {JMLR.org},
booktitle = {Proceedings of the 37th International Conference on Machine Learning},
articleno = {368},
numpages = {10},
series = {ICML'20}
}

@inproceedings{lewis2020rag,
author = {Lewis, Patrick and Perez, Ethan and Piktus, Aleksandra and others},
title = {Retrieval-augmented generation for knowledge-intensive NLP tasks},
year = {2020},
isbn = {9781713829546},
publisher = {Curran Associates Inc.},
address = {Red Hook, NY, USA},
booktitle = {Proceedings of the 34th International Conference on Neural Information Processing Systems},
articleno = {793},
numpages = {16},
location = {Vancouver, BC, Canada},
series = {NIPS '20}
}

@misc{graves2014ntm,
      title={Neural Turing Machines}, 
      author={Alex Graves and Greg Wayne and Ivo Danihelka},
      year={2014},
      eprint={1410.5401},
      archivePrefix={arXiv},
      primaryClass={cs.NE},
      url={https://arxiv.org/abs/1410.5401}, 
}

@misc{raecompressive2019,
      title={Compressive Transformers for Long-Range Sequence Modelling}, 
      author={Jack W. Rae and Anna Potapenko and Siddhant M. Jayakumar and Timothy P. Lillicrap},
      year={2019},
      eprint={1911.05507},
      archivePrefix={arXiv},
      primaryClass={cs.LG},
      url={https://arxiv.org/abs/1911.05507}, 
}

@misc{packer2023memgpt,
      title={MemGPT: Towards LLMs as Operating Systems}, 
      author={Charles Packer and Sarah Wooders and Kevin Lin and Vivian Fang and Shishir G. Patil and Ion Stoica and Joseph E. Gonzalez},
      year={2024},
      eprint={2310.08560},
      archivePrefix={arXiv},
      primaryClass={cs.AI},
      url={https://arxiv.org/abs/2310.08560}, 
}

@inproceedings{zhong2023memorybank,
  title={Memorybank: Enhancing large language models with long-term memory},
  author={Zhong, Wanjun and Guo, Lianghong and Gao, Qiqi and Ye, He and Wang, Yanlin},
  booktitle={Proceedings of the AAAI conference on artificial intelligence},
  volume={38},
  number={17},
  pages={19724--19731},
  year={2024}
}

@misc{chhikara2025mem0,
      title={Mem0: Building Production-Ready AI Agents with Scalable Long-Term Memory}, 
      author={Prateek Chhikara and Dev Khant and Saket Aryan and Taranjeet Singh and Deshraj Yadav},
      year={2025},
      eprint={2504.19413},
      archivePrefix={arXiv},
      primaryClass={cs.CL},
      url={https://arxiv.org/abs/2504.19413}, 
}

@InProceedings{wang2024memoryllmselfupdatablelargelanguage,
  title = 	 {{MEMORYLLM}: Towards Self-Updatable Large Language Models},
  author =       {Wang, Yu and Gao, Yifan and Chen, Xiusi and Jiang, Haoming and Li, Shiyang and Yang, Jingfeng and Yin, Qingyu and Li, Zheng and Li, Xian and Yin, Bing and Shang, Jingbo and Mcauley, Julian},
  booktitle = 	 {Proceedings of the 41st International Conference on Machine Learning},
  pages = 	 {50453--50466},
  year = 	 {2024},
  volume = 	 {235},
  series = 	 {Proceedings of Machine Learning Research},
  month = 	 {21--27 Jul},
  publisher =    {PMLR},

}

@misc{yu2025memagent,
      title={MemAgent: Reshaping Long-Context LLM with Multi-Conv RL-based Memory Agent}, 
      author={Hongli Yu and Tinghong Chen and Jiangtao Feng and Jiangjie Chen and Weinan Dai and Qiying Yu and Ya-Qin Zhang and Wei-Ying Ma and Jingjing Liu and Mingxuan Wang and Hao Zhou},
      year={2026},
      eprint={2507.02259},
      archivePrefix={arXiv},
      primaryClass={cs.CL},
      url={https://arxiv.org/abs/2507.02259}, 
}

@inproceedings{locomo,
    title = "Evaluating Very Long-Term Conversational Memory of {LLM} Agents",
    author = "Maharana, Adyasha  and
      Lee, Dong-Ho  and
      Tulyakov, Sergey  and
      Bansal, Mohit  and
      Barbieri, Francesco  and
      Fang, Yuwei",
    booktitle = "Proceedings of the 62nd Annual Meeting of the {A}ssociation for {C}omputational {L}inguistics (Volume 1: Long Papers)",
    month = aug,
    year = "2024",
    address = "Bangkok, Thailand",
    publisher = "Association for Computational Linguistics",
    url = "https://aclanthology.org/2024.acl-long.747/",
    doi = "10.18653/v1/2024.acl-long.747",
    pages = "13851--13870",
}

@misc{chan2025mlebenchevaluatingmachinelearning,
      title={MLE-bench: Evaluating Machine Learning Agents on Machine Learning Engineering}, 
      author={Jun Shern Chan and Neil Chowdhury and Oliver Jaffe and James Aung and Dane Sherburn and Evan Mays and Giulio Starace and Kevin Liu and others},
      year={2025},
      eprint={2410.07095},
      archivePrefix={arXiv},
      primaryClass={cs.CL},
      url={https://arxiv.org/abs/2410.07095}, 
}

@inproceedings{yang2025gated,
  title={Gated delta networks: Improving mamba2 with delta rule},
  author={Yang, Songlin and Kautz, Jan and Hatamizadeh, Ali},
  booktitle={International Conference on Learning Representations},
  volume={2025},
  pages={29687--29707},
  year={2025}
}

\appendix



\section{Algorithmic Details}
\label{app:algorithm}
This appendix gives the complete context-parallel implementation of Select-TTT in Algorithm~\ref{alg:select-ttt}.
For each chunk, the algorithm computes a query-conditioned relevance score and
the standard In-Place TTT update in parallel. It then converts the scores into
learning-rate multipliers, aggregates the weighted updates with a causal prefix
sum, and applies only preceding updates when producing each chunk's output.

\section{Implementation Details}
\label{app:implementation}

\subsection{Backbone Architectures}
\label{app:backbone}

We use two open-weight decoder-only Transformers. Table~\ref{app:tab-backbone}
summarizes their architectures. Both use SwiGLU gated MLPs and Rotary
Position Embeddings (RoPE); Qwen3-4B additionally applies YaRN
position interpolation~\cite{yarn} to support the long sequences of
LongCite-45k, while Llama-3.1-8B-Instruct is evaluated in its native context window.

\begin{table}[h]
\centering
\caption{Architectural configurations of the two backbones.}
\label{app:tab-backbone}
\footnotesize
\begin{tabular}{lcc}
\toprule
\textbf{Parameter} & \textbf{Llama-3.1-8B-Instruct} & \textbf{Qwen3-4B} \\
\midrule
Parameters (approx.) & $8$B & $4$B \\
Hidden size $d_{\text{model}}$ & $4{,}096$ & $2{,}560$ \\
Num.\ layers & $32$ & $36$ \\
Attention heads & $32$ & $32$ \\
Key-value heads & $8$ & $8$ \\
FFN intermediate size $d_{\text{ff}}$ & $14{,}336$ & $9{,}728$ \\
Vocabulary size & $128{,}256$ & $151{,}936$ \\
RoPE base $\theta$ & $5 \times 10^{5}$ & $10^{6}$ \\
Max position embeddings & $131{,}072$ & $40{,}960$ \\
RoPE extension & None & YaRN (factor $4$) \\
Attention type & Full & Full \\
\bottomrule
\end{tabular}
\end{table}

\subsection{TTT Module Configuration}
\label{app:ttt-config}

Table~\ref{app:tab-ttt} lists the configuration of the In-Place TTT
module and our Select-TTT extension. The module is inserted into the gated
MLP of every sixth layer ($\{0, 6, 12, 18, 24, 30\}$), and the down-projection $\mathbf{W}_{\mathrm{down}}$ of those
layers is treated as the fast weight.

\algnewcommand\algorithmicinput{\textbf{Input:}}
\algnewcommand\Input {\item [\algorithmicinput]}

\begin{algorithm}[H]
\caption{Select-TTT with Context Parallelism}
\label{alg:select-ttt}
\begin{algorithmic}[1]
\Require Pre-trained weights $\theta$ (incl.\ $\mathbf{W}_{\mathrm{up}}$, $\mathbf{W}_{\mathrm{gate}}$, $\mathbf{W}^{(0)}_{\mathrm{down}}$); Conv1D kernel size $k_{\mathrm{conv}}$; target projection $\mathbf{W}_{\mathrm{target}}$; learning rate $\eta$; LM-head $\mathbf{W}_{\mathrm{lm}}$, $\mathbf{b}_{\mathrm{lm}}$;
\Input Chunk sequence $\{\mathbf{X}_{[i]}\}_{i=1}^{T}$; query tokens $\mathbf{q} = (q_{1}, \dots, q_{Q})$.
\Statex \textit{--- Phase 1: compute relevance score and NTP gradient ---}
\ForAll{$i \in \{1, \dots, T\}$ \textbf{in parallel}}
    \State $\mathbf{H}_{[i]} \leftarrow \mathrm{AttentionBlock}(\mathbf{X}_{[i]};\,\theta)$
    \State $\mathbf{Z}^{\mathrm{lm}}_{[i]} \leftarrow \mathrm{RMSNorm}(\mathbf{H}_{[i]})\,\mathbf{W}_{\mathrm{lm}}^{\top} + \mathbf{b}_{\mathrm{lm}}^{\top}$ \Comment{LM-head forward.}
    \State $\widehat{\CE}_{i,c} \leftarrow \mathrm{CE}\bigl((\mathbf{Z}^{\mathrm{lm}}_{[i]})_c;\,\mathbf{q}\bigr),\ \forall c \in \mathbf{X}_{[i]}$ \Comment{Select: per-token CE.}
    \State $\mathbf{CE}_i \leftarrow \mathrm{Pool}^{(K)}_{c \in \mathbf{X}_{[i]}}\bigl(\widehat{\CE}_{i,c}\bigr)$ \Comment{Select: chunk relevance score.}
    \State $\mathbf{U}_{[i]}, \mathbf{G}_{[i]} \leftarrow \mathbf{H}_{[i]}\mathbf{W}_{\mathrm{up}}^{\top},\; \mathbf{H}_{[i]}\mathbf{W}_{\mathrm{gate}}^{\top}$
    \State $\mathbf{Z}_{[i]} \leftarrow \phi(\mathbf{G}_{[i]}) \odot \mathbf{U}_{[i]}$
    \State $\widehat{\mathbf{V}}_{[i]} \leftarrow \mathrm{Conv1D}_{k_{\mathrm{conv}}}(\mathbf{X}_{[i]})\,\mathbf{W}_{\mathrm{target}}$
    \State $\Delta\mathbf{W}_i \leftarrow \widehat{\mathbf{V}}_{[i]}^{\top}\mathbf{Z}_{[i]}$ \Comment{IP-TTT: compute NTP gradient.}
\EndFor
\Statex \textit{--- Phase 2: query-conditioned multipliers and aggregation ---}
\State $\{\lambda_i\}_{i=1}^{T} \leftarrow \mathrm{Clip}\Bigl(T \cdot \mathrm{softmax}\bigl(-\{\mathbf{CE}_i\}_{i=1}^{T} / \tau\bigr),\; 0,\; M\Bigr)$
\State $\{\mathbf{S}_i\}_{i=1}^{T} \leftarrow \mathrm{Cumsum}\bigl(\{\lambda_i\Delta\mathbf{W}_i\}_{i=1}^{T}\bigr)$; $\mathbf{S}_0 \leftarrow \mathbf{0}$

\Statex \textit{--- Phase 3: apply updates and compute outputs ---}
\ForAll{$i \in \{1, \dots, T\}$ \textbf{in parallel}}
    \State $\mathbf{W}^{(i-1)}_{\mathrm{down}} \leftarrow \mathbf{W}^{(0)}_{\mathrm{down}} + \eta\,\mathbf{S}_{i-1}$ \Comment{Use only preceding chunks.}
    \State $\mathbf{O}_{[i]} \leftarrow \mathbf{Z}_{[i]}\bigl(\mathbf{W}^{(i-1)}_{\mathrm{down}}\bigr)^{\top}$
\EndFor
\State \textbf{At document boundaries:} reset fast weights to $\mathbf{W}^{(0)}_{\mathrm{down}}$.
\end{algorithmic}
\end{algorithm}

\begin{table}[h]
\centering
\caption{Configuration of the TTT module and Select-TTT scoring.}
\label{app:tab-ttt}
\footnotesize
\begin{tabular}{lc}
\toprule
\textbf{Component} & \textbf{Value} \\
\midrule
TTT layers & $\{0, 6, 12, 18, 24, 30\}$ \\
Fast weight & MLP down-projection $\mathbf{W}_{\mathrm{down}}$ \\
Conv1D kernel size & $5$ (causal, depthwise, zero-init) \\
Target projection $\mathbf{W}_{\mathrm{target}}$ & $d_{\text{model}} \times d_{\text{model}}$, sparse-diagonal init \\
Objective & NTP-aligned (negative Frobenius inner product) \\
\midrule
Pooling strategy & top-$k$ mean \\
Pooling ratio $\rho$ (default) & $0.1$ \\
Temperature $\tau$ (default) & $0.1$ \\
Cap $M$ (default) & $10.0$ \\
Block size $B$ (default) & $2{,}048$ \\
\bottomrule
\end{tabular}
\end{table}

\paragraph{Initialization.}
At the start of continual training, the newly introduced TTT components are
initialized so that the fast-weight update
$\Delta \mathbf{W}_{\mathrm{down}} = \widehat{\mathbf{V}}_{[i]}^{\top}
\mathbf{Z}_{[i]}$ is negligible, preserving the pre-trained behaviour.
Concretely, the depthwise Conv1D (kernel size $5$, causal padding, no
bias) is zero-initialized, and the projection $\mathbf{W}_{\mathrm{target}}$
is a sparse diagonal matrix whose diagonal entries are drawn from
$\mathcal{N}(0, \sigma^2)$ with $\sigma$ equal to the model's standard
initializer range. This yields $\Delta \mathbf{W}_{\mathrm{down}}
\approx \mathbf{0}$ at step $0$, so the effective $\mathbf{W}_{\mathrm{down}}$
remains identical to its pre-trained value, and the TTT mechanism emerges
smoothly as training progresses.

\subsection{Dataset Statistics}
\label{app:data}

We use separate datasets for inference-time evaluation and continual
pre-training (CPT). The CPT corpus is filtered by language and context length as described below.

\subsubsection{Inference Benchmarks}
\label{app:infer-data}

\textbf{LongMemEval-S} is a benchmark for
long-term interactive memory in chat assistants. It contains $500$ questions
over timestamped histories of approximately $40$ sessions and $115$K tokens. Its six question types are knowledge update, multi-session,
temporal reasoning, single-session user, single-session assistant, and
single-session preference. We use the public benchmark and
follow its official per-subtask accuracy protocol. \textbf{LoCoMo} evaluates very long-term conversational memory.
The public ten-dialogue release contains $1{,}986$ questions over
19--32 sessions and 369--689 turns per dialogue. Its five query types are
single-hop, multi-hop, temporal, open-domain, and adversarial reasoning. We
use the public dialogues and report overall score by query type.

\begin{table}[h]
\centering
\caption{Inference-time evaluation benchmarks.}
\label{app:tab-infer-data}
\footnotesize
\begin{tabular}{lcc}
\toprule
\textbf{Statistic} & \textbf{LongMemEval-S} & \textbf{LoCoMo} \\
\midrule
Questions & $500$ & $1{,}986$ \\
Context organization & $\approx 40$ sessions & $19$--$32$ sessions \\
Context scale & $\approx 115$K tokens & $369$--$689$ turns/dialogue \\
Subtasks / query types & $6$ subtasks & $5$ query types \\
Purpose & inference evaluation & inference evaluation \\
\bottomrule
\end{tabular}
\end{table}

\subsubsection{CPT Dataset}
\label{app:cpt-data}

\textbf{LongCite-45k}~\cite{zhang2024longcite} is a corpus of approximately
$44{,}600$ long-context QA instances with sentence-level citation annotations,
designed to train models to answer questions and identify supporting context
spans. It was constructed using the Coarse-to-Fine (CoF) pipeline and contains
both Chinese and English examples. To maintain data quality and keep the CPT
corpus consistent with our long-context English evaluation benchmarks, we retain only
English examples whose context lengths are in the $8$K--$65$K token range. We
use this filtered English subset for all CPT experiments.

\begin{table}[h]
\centering
\caption{LongCite-45k subset used for continual pre-training.}
\label{app:tab-cpt-data}
\footnotesize
\begin{tabular}{ll}
\toprule
\textbf{Statistic} & \textbf{Value} \\
\midrule
Original corpus & $\approx 44{,}600$ QA instances \\
Languages & Chinese and English (mixed) \\
Subset used for CPT & English only \\
Context-length filter & $8$K--$65$K tokens \\
Annotations & sentence-level citation spans \\
Purpose & continual pre-training \\
\bottomrule
\end{tabular}
\end{table}

\subsection{Training Hyperparameters}
\label{app:train}

Table~\ref{app:tab-train} lists the continual-training hyperparameters for
each backbone. Both backbones are trained with the same optimizer and
global batch size; the TTT learning rate $\eta$ and the chunk size $C$ are
shared across methods within a backbone, so the only difference between
IP-TTT and Select-TTT is whether the per-chunk multiplier is uniform or
query-conditioned. All experiments are conducted on Nvidia H200 GPUs.

\begin{table}[h]
\centering
\caption{Continual-training hyperparameters (Protocol II).}
\label{app:tab-train}
\footnotesize
\begin{tabular}{lcc}
\toprule
\textbf{Hyperparameter} & \textbf{Llama-3.1-8B-Instruct} & \textbf{Qwen3-4B} \\
\midrule
Optimizer & AdamW & AdamW \\
Peak learning rate & $5 \times 10^{-6}$ & $5 \times 10^{-6}$ \\
Weight decay & $0.1$ & $0.1$ \\
Gradient clipping & $1.0$ & $1.0$ \\
LR decay ratio & 0.90 & 0.90 \\
Global batch size & 8 & 8 \\
Warmup ratio & 2\% & 2\%\\ 
Max.\ sequence length & $65{,}536$ & $65{,}536$ \\
Training steps & up to $5$k & up to $5$k \\
Chunk size $C$ & $2{,}048$ & $2{,}048$ \\
TTT learning rate $\eta$ & $0.3$ & $0.3$ \\
RoPE extension during training & None & YaRN (factor $4$) \\
\bottomrule
\end{tabular}
\end{table}

\subsection{Query Extraction and Evaluation}
\label{app:eval}

For every QA instance, the query tokens $\mathbf{q}$ of
Eq.~\eqref{eq:per-token-ce} are obtained by tokenizing the question text
with the backbone's own tokenizer. On
LongMemEval-S we follow the official evaluation protocol and report
accuracy per subtask; on LoCoMo we report overall score per query type. The
relevance-signal analysis of Section~\ref{sec:analysis} uses the
ground-truth answer-chunk annotations of LongMemEval-S's QA instances.

\section{Additional Results}
\label{app:results}

\subsection{Full Numerical Results of Continual-Training}
\label{app:cpt-results}


Tables~\ref{app:tab-cpt-qwen} and~\ref{app:tab-cpt-llama}
report per-subtask accuracy at each 1k-step checkpoint from 1k to 5k
for CPT, IP-TTT, and Select-TTT on both backbones, complementing the Protocol II results.
All values are percentages.

\begin{table}[H]
\centering
\caption{Continual-training results on LongMemEval-S with Qwen3-4B.
All values are percentages (\%).}
\label{app:tab-cpt-qwen}
\scriptsize
\setlength{\tabcolsep}{2.5pt}
\resizebox{\columnwidth}{!}{%
\begin{tabular}{cl cccccc|c}
\toprule
\multirow{2}{*}{\textbf{Step}} & \multirow{2}{*}{\textbf{Method}}
& \multicolumn{7}{c}{\textbf{LongMemEval-S}} \\
\cmidrule(lr){3-9}
& & \textbf{KU} & \textbf{MS} & \textbf{TR} & \textbf{S-Pref.}
& \textbf{S-Asst.} & \textbf{S-User} & \textbf{Avg.} \\
\midrule
\multirow{3}{*}{1000} & CPT & 43.59 & 20.30 & 24.81 & 3.33 & 76.79 & 34.29 & 32.40 \\
& IP-TTT & 35.90 & 24.06 & 28.57 & 3.33 & 73.21 & 37.14 & \textbf{33.20} \\
\rowcolor[HTML]{F2F2F2}
& \textbf{Select-TTT} & 44.87 & 19.55 & 29.32 & 0.00 & 69.64 & 30.00 & 32.00 \\
\midrule
\multirow{3}{*}{2000} & CPT & 43.59 & 26.32 & 25.56 & 0.00 & 71.43 & 34.29 & 33.40 \\
& IP-TTT & 51.28 & 27.07 & 27.07 & 6.67 & 71.43 & 35.71 & 35.80 \\
\rowcolor[HTML]{F2F2F2}
& \textbf{Select-TTT} & 44.87 & 27.07 & 29.32 & 6.67 & 71.43 & 44.29 & \textbf{36.60} \\
\midrule
\multirow{3}{*}{3000} & CPT & 43.59 & 22.56 & 28.57 & 3.33 & 69.64 & 40.00 & 34.00 \\
& IP-TTT & 42.31 & 24.06 & 27.82 & 3.33 & 67.86 & 42.86 & 34.20 \\
\rowcolor[HTML]{F2F2F2}
& \textbf{Select-TTT} & 50.00 & 22.56 & 25.56 & 3.33 & 75.00 & 40.00 & \textbf{34.80} \\
\midrule
\multirow{3}{*}{4000} & CPT & 43.59 & 19.55 & 33.83 & 0.00 & 71.43 & 47.14 & 35.60 \\
& IP-TTT & 46.15 & 26.32 & 29.32 & 3.33 & 67.86 & 45.71 & 36.20 \\
\rowcolor[HTML]{F2F2F2}
& \textbf{Select-TTT} & 43.59 & 22.56 & 33.08 & 3.33 & 69.64 & 50.00 & \textbf{36.60} \\
\midrule
\multirow{3}{*}{5000} & CPT & 38.46 & 24.06 & 33.08 & 10.00 & 67.86 & 40.00 & 35.00 \\
& IP-TTT & 41.03 & 26.32 & 31.58 & 10.00 & 64.29 & 42.86 & 35.60 \\
\rowcolor[HTML]{F2F2F2}
& \textbf{Select-TTT} & 50.00 & 24.06 & 31.58 & 0.00 & 67.86 & 51.43 & \textbf{37.40} \\
\bottomrule
\end{tabular}%
}
\end{table}

\begin{table}[t]
\centering
\caption{Continual-training results on LongMemEval-S with Llama-3.1-8B-Instruct.
All values are percentages (\%). Avg. is the question-weighted overall accuracy.
Bold Avg. values indicate the best result at each training step.}
\label{app:tab-cpt-llama}
\scriptsize
\setlength{\tabcolsep}{2.5pt}
\resizebox{\columnwidth}{!}{%
\begin{tabular}{cl cccccc|c}
\toprule
\multirow{2}{*}{\textbf{Step}} & \multirow{2}{*}{\textbf{Method}}
& \multicolumn{7}{c}{\textbf{LongMemEval-S}} \\
\cmidrule(lr){3-9}
& & \textbf{KU} & \textbf{MS} & \textbf{TR} & \textbf{S-Pref.}
& \textbf{S-Asst.} & \textbf{S-User} & \textbf{Avg.} \\
\midrule
\multirow{3}{*}{1000} & CPT & 29.49 & 22.56 & 30.08 & 0.00 & 71.43 & 35.71 & 31.60 \\
& IP-TTT & 44.87 & 18.80 & 24.81 & 6.67 & 71.43 & 32.86 & 31.60 \\
\rowcolor[HTML]{F2F2F2}
& \textbf{Select-TTT} & 43.59 & 21.05 & 25.56 & 3.33 & 76.79 & 35.71 & \textbf{33.00} \\
\midrule
\multirow{3}{*}{2000} & CPT & 44.87 & 20.30 & 25.56 & 3.33 & 73.21 & 37.14 & 32.80 \\
& IP-TTT & 52.56 & 24.81 & 32.33 & 0.00 & 69.64 & 37.14 & \textbf{36.40} \\
\rowcolor[HTML]{F2F2F2}
& \textbf{Select-TTT} & 46.15 & 21.80 & 29.32 & 6.67 & 71.43 & 44.29 & 35.40  \\
\midrule
\multirow{3}{*}{3000} & CPT & 43.59 & 21.80 & 27.82 & 3.33 & 73.21 & 41.43 & 34.20 \\
& IP-TTT & 44.87 & 21.80 & 28.57 & 6.67 & 71.43 & 45.71 & 35.20 \\
\rowcolor[HTML]{F2F2F2}
& \textbf{Select-TTT} & 50.00 & 24.81 & 28.57 & 3.33 & 69.64 & 47.14 & \textbf{36.60} \\
\midrule
\multirow{3}{*}{4000} & CPT & 46.15 & 22.56 & 24.06 & 0.00 & 64.29 & 48.57 & 33.60 \\
& IP-TTT & 43.59 & 20.30 & 26.32 & 3.33 & 71.43 & 47.14 & 34.00 \\
\rowcolor[HTML]{F2F2F2}
& \textbf{Select-TTT} & 51.28 & 25.56 & 24.06 & 3.33 & 73.21 & 51.43 & \textbf{36.80} \\
\midrule
\multirow{3}{*}{5000} & CPT & 41.03 & 23.31 & 33.08 & 6.67 & 67.86 & 42.86 & 35.40 \\
& IP-TTT & 47.44 & 24.81 & 29.32 & 3.33 & 73.21 & 45.71 & 36.60 \\
\rowcolor[HTML]{F2F2F2}
& \textbf{Select-TTT} & 56.41 & 26.32 & 34.59 & 3.33 & 71.43 & 47.14 & \textbf{39.80} \\
\bottomrule
\end{tabular}%
}
\end{table}




\subsection{Full Numerical Results of the Ablation Study}
\label{app:ablation-full}

The following six tables provide per-subtask LongMemEval-S results for the
temperature ($\tau$), pooling-ratio ($\rho$), and pooling-strategy
ablations on both backbones.

\paragraph{Temperature $\tau$.}
Tables~\ref{app:tab-ablation-temp-llama} and~\ref{app:tab-ablation-temp-qwen}
report the per-subtask results for $\tau\in\{0.1,0.3,0.5\}$ on the two
backbones. All values are percentages.

\begin{table}[H]
\centering
\caption{Temperature ablation on LongMemEval-S with Llama-3.1-8B-Instruct.}
\label{app:tab-ablation-temp-llama}
\scriptsize
\setlength{\tabcolsep}{2.5pt}
\resizebox{\columnwidth}{!}{%
\begin{tabular}{cc cccccc|c}
\toprule
\multirow{2}{*}{\textbf{Method}} & \multirow{2}{*}{\textbf{$\tau$}} & \multicolumn{7}{c}{\textbf{LongMemEval-S}} \\
\cmidrule(lr){3-9}
& & \textbf{KU} & \textbf{MS} & \textbf{TR} & \textbf{S-Pref.} & \textbf{S-Asst.} & \textbf{S-User} & \textbf{Avg.} \\
\midrule
\multirow{3}{*}{\textbf{Select-TTT}} & 0.1 & 52.56 & 22.56 & 28.57 & 13.33 & 64.29 & 54.29 & 37.40 \\
 & 0.3 & 46.15 & 21.05 & 23.31 & 3.33 & 66.07 & 57.14 & 34.60 \\
 & 0.5 & 46.15 & 20.30 & 19.55 & 10.00 & 64.29 & 57.14 & 33.60 \\
\bottomrule
\end{tabular}%
}
\end{table}

\begin{table}[H]
\centering
\caption{Temperature ablation on LongMemEval-S with Qwen3-4B.}
\label{app:tab-ablation-temp-qwen}
\scriptsize
\setlength{\tabcolsep}{2.5pt}
\resizebox{\columnwidth}{!}{%
\begin{tabular}{cc cccccc|c}
\toprule
\multirow{2}{*}{\textbf{Method}} & \multirow{2}{*}{\textbf{$\tau$}} & \multicolumn{7}{c}{\textbf{LongMemEval-S}} \\
\cmidrule(lr){3-9}
& & \textbf{KU} & \textbf{MS} & \textbf{TR} & \textbf{S-Pref.} & \textbf{S-Asst.} & \textbf{S-User} & \textbf{Avg.} \\
\midrule
\multirow{3}{*}{\textbf{Select-TTT}} & 0.1 & 29.49 & 26.32 & 22.56 & 3.33 & 57.14 & 32.86 & 28.80 \\
 & 0.3 & 33.33 & 22.56 & 21.05 & 0.00 & 60.71 & 25.71 & 27.20 \\
 & 0.5 & 28.21 & 21.05 & 21.05 & 0.00 & 60.71 & 27.14 & 26.20 \\
\bottomrule
\end{tabular}%
}
\end{table}

\paragraph{Pooling ratio $\rho$.}
Tables~\ref{app:tab-ablation-ratio-llama} and~\ref{app:tab-ablation-ratio-qwen}
report the per-subtask results for $\rho\in\{0.01,0.05,0.10,0.20\}$. All values are percentages.

\begin{table}[H]
\centering
\caption{Pooling-ratio ablation on LongMemEval-S with Llama-3.1-8B-Instruct.}
\label{app:tab-ablation-ratio-llama}
\scriptsize
\setlength{\tabcolsep}{2.5pt}
\resizebox{\columnwidth}{!}{%
\begin{tabular}{cc cccccc|c}
\toprule
\multirow{2}{*}{\textbf{Method}} & \multirow{2}{*}{\textbf{$\rho$}} & \multicolumn{7}{c}{\textbf{LongMemEval-S}} \\
\cmidrule(lr){3-9}
& & \textbf{KU} & \textbf{MS} & \textbf{TR} & \textbf{S-Pref.} & \textbf{S-Asst.} & \textbf{S-User} & \textbf{Avg.} \\
\midrule
\multirow{4}{*}{\textbf{Select-TTT}} & 0.01 & 53.85 & 23.31 & 24.06 & 6.67 & 64.29 & 54.29 & 36.20 \\
 & 0.05 & 53.85 & 21.05 & 25.56 & 13.33 & 60.71 & 54.29 & 36.00 \\
 & 0.10 & 52.56 & 22.56 & 28.57 & 13.33 & 64.29 & 54.29 & 37.40 \\
 & 0.20 & 53.85 & 22.56 & 25.56 & 13.33 & 66.07 & 50.00 & 36.40 \\
\bottomrule
\end{tabular}%
}
\end{table}

\begin{table}[H]
\centering
\caption{Pooling-ratio ablation on LongMemEval-S with Qwen3-4B.}
\label{app:tab-ablation-ratio-qwen}
\scriptsize
\setlength{\tabcolsep}{2.5pt}
\resizebox{\columnwidth}{!}{%
\begin{tabular}{cc cccccc|c}
\toprule
\multirow{2}{*}{\textbf{Method}} & \multirow{2}{*}{\textbf{$\rho$}} & \multicolumn{7}{c}{\textbf{LongMemEval-S}} \\
\cmidrule(lr){3-9}
& & \textbf{KU} & \textbf{MS} & \textbf{TR} & \textbf{S-Pref.} & \textbf{S-Asst.} & \textbf{S-User} & \textbf{Avg.} \\
\midrule
\multirow{4}{*}{\textbf{Select-TTT}} & 0.01 & 37.18 & 18.80 & 20.30 & 3.33 & 64.29 & 21.43 & 26.60 \\
 & 0.05 & 32.05 & 18.05 & 20.30 & 3.33 & 57.14 & 32.86 & 26.40 \\
 & 0.10 & 29.49 & 26.32 & 22.56 & 3.33 & 57.14 & 32.86 & 28.80 \\
 & 0.20 & 35.90 & 20.30 & 18.80 & 3.33 & 60.71 & 35.71 & 28.00 \\
\bottomrule
\end{tabular}%
}
\end{table}

\paragraph{Pooling strategy.}
Tables~\ref{app:tab-ablation-pool-llama} and~\ref{app:tab-ablation-pool-qwen}
compare Max, Mean, and top-$k$ mean pooling on both backbones.
All values are percentages.

\begin{table}[H]
\centering
\caption{Pooling-strategy ablation on LongMemEval-S with Llama-3.1-8B-Instruct.}
\label{app:tab-ablation-pool-llama}
\scriptsize
\setlength{\tabcolsep}{2.5pt}
\resizebox{\columnwidth}{!}{%
\begin{tabular}{cc cccccc|c}
\toprule
\multirow{2}{*}{\textbf{Method}} & \multirow{2}{*}{\textbf{Pooling}} & \multicolumn{7}{c}{\textbf{LongMemEval-S}} \\
\cmidrule(lr){3-9}
& & \textbf{KU} & \textbf{MS} & \textbf{TR} & \textbf{S-Pref.} & \textbf{S-Asst.} & \textbf{S-User} & \textbf{Avg.} \\
\midrule
\multirow{3}{*}{\textbf{Select-TTT}} & Max & 53.85 & 21.80 & 18.80 & 6.67 & 67.86 & 60.00 & 35.60 \\
 & Mean & 50.00 & 19.55 & 20.30 & 3.33 & 64.29 & 45.71 & 32.20 \\
 & Top-$k$ mean & 52.56 & 22.56 & 28.57 & 13.33 & 64.29 & 54.29 & 37.40 \\
\bottomrule
\end{tabular}%
}
\end{table}

\begin{table}[H]
\centering
\caption{Pooling-strategy ablation on LongMemEval-S with Qwen3-4B.}
\label{app:tab-ablation-pool-qwen}
\scriptsize
\setlength{\tabcolsep}{2.5pt}
\resizebox{\columnwidth}{!}{%
\begin{tabular}{cc cccccc|c}
\toprule
\multirow{2}{*}{\textbf{Method}} & \multirow{2}{*}{\textbf{Pooling}} & \multicolumn{7}{c}{\textbf{LongMemEval-S}} \\
\cmidrule(lr){3-9}
& & \textbf{KU} & \textbf{MS} & \textbf{TR} & \textbf{S-Pref.} & \textbf{S-Asst.} & \textbf{S-User} & \textbf{Avg.} \\
\midrule
\multirow{3}{*}{\textbf{Select-TTT}} & Max & 38.46 & 20.30 & 18.80 & 3.33 & 60.71 & 35.71 & 28.40 \\
 & Mean & 33.33 & 20.30 & 21.05 & 0.00 & 62.50 & 28.57 & 27.20 \\
 & Top-$k$ mean & 29.49 & 26.32 & 22.56 & 3.33 & 57.14 & 32.86 & 28.80 \\
\bottomrule
\end{tabular}%
}
\end{table}

\section{Online Resources}

The full source code, training and evaluation configurations, and the
Select-TTT-enhanced checkpoints will be released at the project repository
upon publication.

\section{Usage of LLMs}
\label{app:llm-usage}

Large language models are used in this work in two capacities. First, the
method itself operates on open-weight LLMs (Llama-3.1-8B-Instruct and
Qwen3-4B). Second, writing
assistance was limited to grammar and phrasing suggestions; all technical
content, experiments, and final text were prepared and verified by the
authors, who take full responsibility for the paper.

\end{document}